\documentclass[twocolumn,fleqn]{article}

\usepackage{arxiv}

\usepackage[utf8]{inputenc}
\usepackage[T1]{fontenc}
\usepackage[colorlinks]{hyperref}
\usepackage{url}
\usepackage{amsmath,amsfonts,amssymb}
\usepackage{booktabs,graphicx}
\usepackage{makecell,multirow,array,colortbl}
\usepackage[table,svgnames,dvipsnames]{xcolor}
\usepackage{tabularx}
\usepackage{xspace}
\usepackage{microtype}
\usepackage[numbers]{natbib}
\usepackage{cleveref}

\AtBeginDocument{\newgeometry{textheight=9in,textwidth=7in,top=1in,headheight=14pt,headsep=25pt,footskip=30pt}}
\makeatletter
\renewcommand{\maketitle}{%
  \par
  \begingroup
    \renewcommand{\thefootnote}{\fnsymbol{footnote}}
    \long\def\@makefntext##1{%
      \parindent 1em\noindent
      \hbox to 1.8em{\hss $\m@th ^{\@thefnmark}$}##1
    }
    \twocolumn[\@maketitle]
    \thispagestyle{empty}
    \@thanks
  \endgroup
  \let\maketitle\relax
  \let\thanks\relax
}
\makeatother
\renewenvironment{abstract}
{\centerline{\large \bfseries \scshape Abstract}\par\smallskip}
{\par}

\newcommand{\gram}{\textsc{Gram}\xspace}
\newcommand{\hypergram}{\textsc{HyperGram}\xspace}
\newcommand{\vast}{\textsc{VAST}\xspace}
\newcommand{\tri}{\textsc{Triangle}\xspace}
\newcommand{\pmrl}{\textsc{PMRL}\xspace}
\newcommand{\hy}{\textsc{HyVol}\xspace}
\newcommand{\why}{\textsc{wHyVol}\xspace}
\newcommand{\wrhg}{\textsc{wRHG}\xspace}

\title{Fuse To Align: Multimodal Representation Learning for Retrieval via Hypergraph Regularization}

\author{%
	Anindya Nag$^{1}$\thanks{These authors contributed equally to this work.} \quad
	Ambuj Mehrish$^{1}$\footnotemark[1] \quad
	Sebastiano Vascon$^{1,2}$\thanks{Corresponding author.} \\
	\normalfont $^{1}$Department of Environmental Sciences, Informatics and Statistics \\
	\normalfont Ca' Foscari University of Venice, Italy \\
	\normalfont $^{2}$European Center for Living Technology, Venice, Italy \\
	\normalfont \small \texttt{\{anindya.nag, ambuj.mehrish, sebastiano.vascon\}@unive.it}
}

\date{}
\renewcommand{\headeright}{}
\renewcommand{\undertitle}{}

\renewcommand{\shorttitle}{Multimodal Representation Learning for Retrieval via Hypergraph Regularization}

\hypersetup{
linkcolor=DarkSlateGrey, urlcolor=DarkSlateGrey, citecolor=DarkSlateGrey, filecolor=DarkSlateGrey,
pdftitle={Fuse To Align: Multimodal Representation Learning for Retrieval via Hypergraph Regularization},
pdfauthor={Anindya Nag, Ambuj Mehrish, Sebastiano Vascon},
pdfkeywords={Multimodal Retrieval, Information Fusion, Hypergraph Learning, Gramian Volume, Missing Modalities},
}

\begin{document}
\maketitle

\begin{abstract}
Multimodal retrieval must integrate complementary evidence from views that may be missing, noisy, or unequally informative. When a sample provides limited reliable evidence, learning from its views alone can hinder the acquisition of useful retrieval representations. Semantically related training samples may provide complementary observations, yet existing volume-based objectives jointly align modalities within each sample without explicitly enabling information exchange across samples. We therefore propose a training-time hypergraph fusion framework that connects observed modalities both within samples and across semantically related samples, allowing representation learning to draw on evidence beyond an individual sample while leaving the retrieval pipeline unchanged at inference. Observed modalities form vertices, while caption-guided semantic neighbourhoods establish cross-sample connections. The framework includes HyVol as a special case and introduces two configurations: wHyVol learns view contribution weights and symmetric semantic affinities, whereas wRHG additionally models typed interactions and participant roles through relational propagation. Missing modalities are omitted rather than imputed, and the hypergraph is discarded after training, incurring no additional inference cost. Experiments on eleven zero-shot video--text and audio--text retrieval benchmarks demonstrate broad improvements with both GRAM and HyperGRAM. Across 17 video benchmark--modality settings evaluated in both retrieval directions, wRHG outperforms GRAM in all 34 comparisons and HyperGRAM in 33 of 34. Mean text-to-video R@1 increases from 63.4 to 68.2 with GRAM and from 64.8 to 68.4 with HyperGRAM, while the largest video-to-text gain reaches 13.3 percentage points on ActivityNet. Under missing-modality evaluation, both weighted configurations retain positive text-to-video gains over GRAM across seven benchmarks at every evaluated missingness rate, even when 90\% of clips lack one modality. These results support training-time semantic exchange as an effective mechanism for learning retrieval representations from complementary multi-view evidence while preserving the original inference procedure.
\end{abstract}


\section{Introduction}
\label{sec:intro}

\begin{figure*}[t]
\centering
\includegraphics[width=0.8\textwidth]{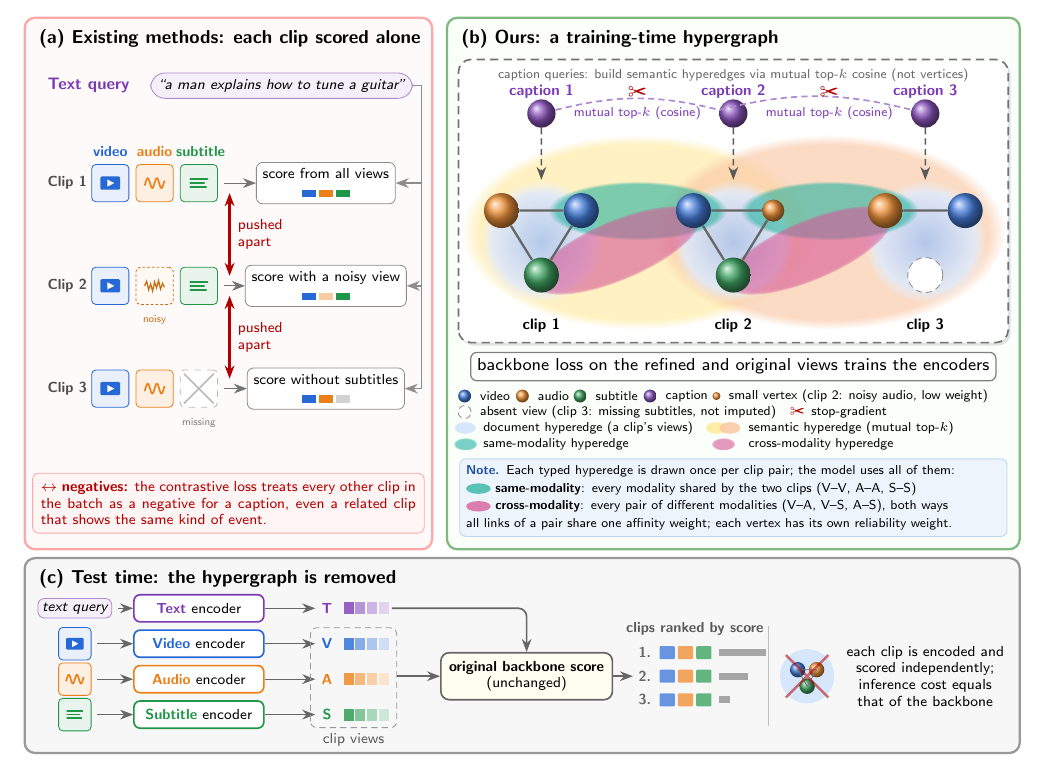}
\caption{Overview of the proposed framework. (a) Existing methods score each clip alone: a clip's score depends only on the caption and the clip's own views, and every other clip in the batch, even a related one, is pushed apart as a negative. (b) Our training-time hypergraph: observed views form vertices, document hyperedges group the views of each clip, and caption-guided semantic hyperedges connect related clips through same-modality and cross-modality links; missing views have no vertex and are not imputed, and each vertex receives a learned reliability weight. (c) At test time the hypergraph is removed: each clip is encoded and scored independently with the original backbone score, so the inference cost equals that of the backbone.}
\label{fig:overview}
\end{figure*}

Humans interpret the world by integrating information across multiple sensory modalities. For example, visual input can influence auditory perception, while language provides contextual cues that help unify sensory experiences. Psychologists refer to this process as multisensory integration, through which individuals reconstruct events from both overlapping and modality-specific sensory cues. Multimodal representation learning presents an analogous challenge because images, sounds, speech, and text capture complementary aspects of the same scenario. Effective representation learning must therefore identify shared features across modalities while preserving the distinctive information provided by each modality.

Text-to-video retrieval requires a model to identify, among thousands of clips, the clip described by a given caption. Each clip provides multiple views of the same event, including frames, a soundtrack, and subtitles, but real-world collections rarely contain all views at consistently high quality. Consider two clips described as ``a man explains how to tune a guitar''. In the first clip, the frames depict the guitar, and the subtitles transcribe the explanation. In the second, the subtitles are absent, and the soundtrack contains only background music, leaving the frames as the sole reliable source of evidence. Such missing, noisy, and unequally informative views are common, requiring a retrieval model to integrate them into a single relevance score.

Current multimodal retrieval models evaluate each clip independently. Contrastive foundation models such as CLIP \cite{Radford2021LearningTV} measure the cosine similarity between two embeddings, while their multi-view extensions either align each view with an anchor modality or fuse multiple views into a single representation~\cite{Girdhar2023ImageBindOE,zhu2024languagebind,chen2023vast,liu2024valor}. Geometric objectives extend this approach by jointly scoring all views of a clip using the volume spanned with the caption in \gram~\cite{cicchetti2025gramian}, the area of a triangle in TRIANGLE~\cite{cicchetti2026triangle}, the dominant singular value in PMRL~\cite{liu2026principled}, or a combination of Euclidean and hyperbolic volumes in \hypergram~\cite{na2026hypergram}. Nevertheless, in all these methods, a clip's score depends exclusively on the caption and the clip's own views. Consequently, in the example above, the second clip must be matched using only one reliable view, while the first clip, despite depicting the same event, is treated solely as a negative during training~\cite{ma2021smil,ma2022multimodal}.

Our central premise is that fusion should exploit complementary evidence both within a sample and across semantically related training samples. When a clip contains missing or uninformative views, restricting fusion to its own observations limits the evidence available for representation learning. Related clips may contain useful observations of similar events, providing an additional source of supervision through information exchange. This motivates a fusion mechanism that jointly models cross-modal dependencies within clips and semantic associations across clips.

We propose a training-time hypergraph feature fusion framework that captures both forms of interaction. Each observed candidate modality forms a vertex, while hyperedges connect modalities belonging to the same candidate and observations from semantically related candidates. Feature propagation along these hyperedges integrates contextual evidence into the modality-specific representations, which remain separate inputs to the retrieval objective. Learned contribution weights regulate this fusion, while relational propagation distinguishes interaction types and participant roles. Missing views remain absent rather than being reconstructed. The hypergraph operates only during training, allowing the encoders to learn from fused evidence while preserving independent candidate encoding and the original retrieval procedure at inference. Figure~\ref{fig:overview} contrasts this design with existing methods, which score each clip alone, and shows that the hypergraph is removed at test time.

Our framework differs from conventional fusion strategies in three respects. First, it integrates evidence both within individual clips and across semantically related training clips. Second, unlike approaches that reconstruct missing views~\cite{ma2021smil}, it operates only on observed modalities, enabling information exchange without imputation. Third, fusion is confined to training: the learned encoders retain their benefits, while inference requires neither a hypergraph nor interactions among evaluation candidates. The framework encompasses HyVol~\cite{nag2027hyvol} as a special case, where document and semantic neighborhood hyperedges support additive feature fusion. Our weighted and relational configurations investigate how to regulate individual-view contributions and related-sample contributions, and how different interactions should participate in the fusion.

HyVol does not explicitly model differences in view reliability or distinguish modality-specific interaction types. A silent soundtrack receives the same weight as the frames of the corresponding clip; related clips are weighted by an attention score that ranks the neighbors of every clip in the same order~\cite{brody2022attentive}; and a link between the frames of two clips is not distinguished from a link between the frames of one clip and the subtitles of another. Moreover, only the refined embeddings are trained, although retrieval subsequently uses the unrefined embeddings. Information exchange across clips therefore raises two questions that remain unresolved in HyVol: how much should each view and each related clip contribute, and what type of relationship connects them?

\textbf{Weighted HyVol (wHyVol)} addresses the first question through two learned weights. The first is a view-specific reliability weight predicted solely from the corresponding view embedding. In the guitar example, the background music in the second clip may receive a low weight, while its frames receive a high weight, causing the clip representation to be determined primarily by reliable evidence. The reliability weight therefore provides an adaptive fusion mechanism for imbalanced views: each view's contribution depends on its estimated quality rather than its modality. The second weight represents the affinity between two related clips; it is computed jointly from their captions and is symmetric. Consequently, a clip receives substantial information from a close semantic match but little information from a weak or incidental match. Unlike the attention mechanism in HyVol, this affinity can induce different neighbour rankings for different clips.

The \textbf{weighted relational hypergraph (wRHG)} addresses the second question. It retains both weights introduced by wHyVol but replaces untyped links with typed relations. A same-modality link connects corresponding views from two related clips, such as their frames; a cross-modality link connects different views, such as the frames of one clip and the subtitles of another; and a whole-clip link connects all observed views from both clips. Each relation type has distinct learned parameters, and the message received by a view through a link depends jointly on all other views associated with that link. Thus, a clip with missing subtitles can learn from a related clip's subtitles through a cross-modality link, while the model learns the relative importance of each relation type.

In \why and \wrhg, a missing view has no corresponding vertex in the hypergraph; consequently, neither model imputes nor replaces missing views. These configurations also train the unrefined embeddings used at test time. As with \hy, their hypergraph modules operate exclusively during training. We integrate \hy, \why, and \wrhg with two volume-based backbones, \gram and \hypergram, which share the pretrained vision--audio--language foundation encoders (VAST~\cite{chen2023vast}). We evaluate all configurations zero-shot on eleven text--video and text--audio retrieval benchmarks. Averaged across the $17$ text-to-video settings from the eight video benchmarks, the R@1 of \gram increases from $63.4$ to $68.2$ with wRHG (and to $67.3$ with \hy), while that of \hypergram increases from $64.8$ to $68.4$. The gains are consistent across datasets: \wrhg improves the R@1 of \gram in all $17$ text-to-video and all $17$ video-to-text settings, and improves that of \hypergram in $16$ and $17$ settings, respectively. The weighted variants also retain gains under missing-modality evaluation, with consistent improvements over GRAM and benchmark-dependent results with \hypergram. Our main contributions are as follows.
\begin{itemize}
\item We formulate multi-view feature fusion within and across semantically related samples through a common training-time hypergraph framework, with \hy recovered as a special case.
\item We introduce two new instances of this framework: a weighted hypergraph (\why), which learns the contribution of each view and related clip, and a weighted relational hypergraph (\wrhg), which models distinct interaction types through relation-specific parameters and participant roles. Both methods handle missing views without imputation and incur no additional inference cost.
\item We evaluate the proposed methods using two backbones and eleven zero-shot benchmarks---eight for text--video retrieval and three for text--audio retrieval---and conduct a missing-modality study across seven video benchmarks.

\end{itemize}

Building on HyVol~\cite{nag2027hyvol}, this article introduces a general framework together with wHyVol and wRHG, expands the evaluation from six to eleven benchmarks, and extends the missing-modality study from four to seven video benchmarks. The remainder of this article is organized as follows. Section~\ref{sec:related} reviews the related literature. Section~\ref{sec:method} presents the general hypergraph framework, recovers HyVol as a special case, and introduces wHyVol and wRHG. Section~\ref{sec:Experiments} describes the datasets and evaluation protocol and reports the zero-shot and missing-modality results. Section~\ref{sec:Discussion} discusses these results, Section~\ref{sec:Limitations} outlines the limitations of the approach, and Section~\ref{sec:conclusion} concludes the article.

\section{Related Work}
\label{sec:related}
Our research builds upon multimodal representation learning, higher-order geometric alignment, and hypergraph-based information fusion. We examine these foundational concepts alongside methods that address missing and imbalanced modalities, inter-sample structural relationships in retrieval tasks, and relation-aware message passing.

\subsection{Contrastive Vision--Language and Multimodal Foundations}
CLIP~\cite{Radford2021LearningTV} established contrastive pretraining on image--text pairs, whereby representations of matched pairs are brought closer together while all other pairs in the batch serve as negatives. Subsequent work developed larger and openly available backbones~\cite{Sun2023EVACLIPIT}. 
More recent approaches have developed large-scale video--language models pretrained on extensive video corpora~\cite{Zeng2023TVTSv2LO,Ye2022HiTeAHT,Wang2022InternVideoGV,Xu2023mPLUG2AM,Wang2022OmniVLOF,Zhao2024VideoPrismAF}. 
Contrastive learning has also been applied to audio--text retrieval~\cite{Nagrani2022LearningAM,Zhao2021ConnectingTD} and to text--video retrieval through multigranular alignment and text-conditioned objectives. 

When more than two modalities are available, existing methods typically either fuse streams into a single representation or align them within a shared embedding space. VALOR~\cite{liu2024valor} fuses video and audio, while VAST~\cite{chen2023vast} additionally incorporates subtitles. ImageBind~\cite{Girdhar2023ImageBindOE} and LanguageBind~\cite{zhu2024languagebind} instead align modalities using images and language as anchors, respectively. Alignment may also precede explicit cross-modal fusion~\cite{Li2021AlignBF}. These approaches do not directly score joint agreement among a candidate's individual streams. Our models retain the modality-specific VAST embeddings and evaluate their joint agreement with the query.

\subsection{Higher-Order Geometric Alignment}
Recent methods assess multimodal agreement through joint geometric objectives. GRAM~\cite{cicchetti2025gramian} uses the Gramian volume spanned by the query and candidate modality embeddings, while \tri~\cite{cicchetti2026triangle} measures the triangle area formed by text, video, and audio embeddings. \pmrl~\cite{liu2026principled} promotes rank-one alignment by maximising the dominant singular value of the embedding matrix, and \hypergram~\cite{na2026hypergram} combines Euclidean and hyperbolic volumes. These methods score each candidate using only its own embeddings. \hy~\cite{nag2027hyvol} introduces information exchange across candidates through a training-time hypergraph while retaining the \gram and \hypergram scoring functions. Our framework includes \hy as a special case and investigates learned contribution weights, semantic affinities, and typed relational propagation.

\subsection{Learning with Missing and Imbalanced Modalities}
Real-world collections often contain missing modalities and streams that differ in informativeness. Existing methods address missing views through reconstruction~\cite{ma2021smil} or fusion mechanisms designed to accommodate absent inputs~\cite{ma2022multimodal}. Our framework instead excludes missing modalities from the hypergraph and the Gram matrix, so retrieval depends only on observed streams. Learned vertex weights predicted from individual modality embeddings regulate their contributions during training-time fusion.

\subsection{Hypergraphs and Inter-Sample Structure in Retrieval}
Hypergraphs represent higher-order relations by connecting multiple vertices within a single hyperedge. Hypergraph neural networks exchange features between vertices and hyperedges~\cite{feng2019hypergraph}, while AllSet~\cite{chien2022allset} learns set-based aggregation functions. Applications include cross-modal retrieval through bipartite hypergraphs~\cite{li2024educross}. Attention mechanisms learn interaction weights~\cite{velickovic2018graph}, although the original additive attention score has a static neighbor-ranking property~\cite{brody2022attentive}, motivating the pair-dependent affinity kernel in wHyVol. In retrieval, inter-item structure is also exploited through test-time methods such as $k$-reciprocal re-ranking~\cite{zhong2017reranking} and querybank normalisation~\cite{bogolin2022qbnorm}. By contrast, HyVol~\cite{nag2027hyvol} and our configurations exchange information only during training, allowing independent candidate encoding without additional inference cost.

\subsection{Relational and Knowledge Hypergraphs}
Relational graph convolutional networks~\cite{schlichtkrull2018rgcn} assign distinct transformations to different edge types. Knowledge hypergraphs extend this principle to facts involving more than two arguments, where the position of each entity carries semantic information. HypE~\cite{fatemi2020hype} embeds entities using position-specific filters; StarE~\cite{galkin2020stare} represents hyper-relational facts with qualifiers; and G-MPNN~\cite{yadati2020gmpnn} performs message passing over multi-relational ordered hypergraphs. HC-MPNN~\cite{huang2024hcmpnn} combines relation embeddings with positional encodings and constructs the message to each member of a fact from all remaining members. Its expressive power for link prediction has been characterised for relations of arbitrary arity.

Graphs with multiple edge types have also been applied to multimodal fusion. For conversational emotion recognition, MMGCN~\cite{hu2021mmgcn} connects utterances through modality-specific and cross-modal edges. For unaligned multimodal language sequences, MTAG~\cite{yang2021mtag} constructs a graph whose edges depend on node modalities and temporal order. In each setting, the data determine the edges or facts, such as a knowledge base, conversational order, or temporal order. In wRHG, by contrast, facts are constructed within each training batch according to caption similarity, and each fact's relation type identifies the modalities and candidates that it connects.

\section{Relational Hypergraph Learning}
\label{sec:method}

Multimodal representation learning must capture dependencies at two complementary levels: cross-modal associations within each candidate and semantic correspondences across candidates. The former integrates complementary evidence from the modalities observed for a candidate, while the latter lets its representation incorporate relevant information from semantically related candidates in the same training batch. We model both forms of dependency using a weighted hypergraph in which each vertex represents an observed candidate--modality pair. Hyperedge construction determines which vertices interact, learned weights control their contributions, and the propagation operator specifies how the resulting information is aggregated.

\begin{figure*}[t]
\centering
\includegraphics[width=0.95\textwidth]{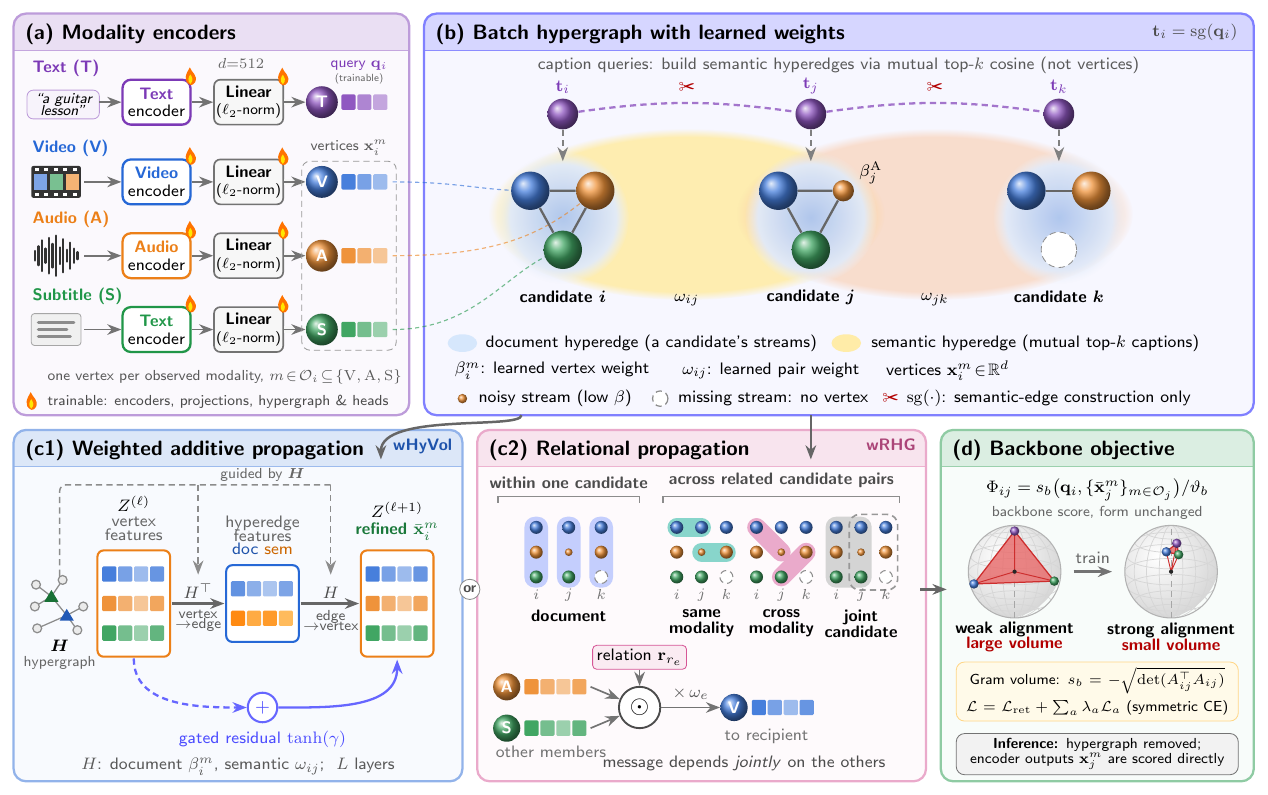}
\caption{\textbf{Overview of training-time hypergraph refinement.} Each observed modality becomes a weighted vertex, and stop-gradient captions connect semantically related candidates through mutual top-$k$ neighbors. Information is then exchanged either by weighted additive propagation (\textsc{\why}) or by typed relational messages (\textsc{wRHG}), and the refined embeddings are scored by the unchanged \textsc{\gram}/\textsc{\hypergram} objective. The hypergraph is used only during training and is removed at inference.}

\label{fig:method}
\end{figure*}

The framework supports two propagation mechanisms: weighted additive aggregation and relation-conditioned multiplicative interaction. Both mechanisms refine candidate embeddings under the original retrieval objective without imposing a specific similarity function or representation geometry. \hy~\cite{nag2027hyvol} is recovered as a special case by choosing a specific hyperedge construction, weighting scheme, and additive propagation operator. The hypergraph is used only during training; at inference, the model retains the trained encoders and follows the original retrieval procedure without constructing or propagating the hypergraph. Figure~\ref{fig:method} summarises the framework.

\subsection{Problem Formulation}
\label{sec:problem}

Let a mini-batch comprise $N$ matched query--item pairs $\{(c_i,y_i)\}_{i=1}^{N}$, where $c_i$ is a caption query and $y_i$ is its matched retrieval item. We use \emph{candidate} to denote any retrieval item evaluated as a potential match for a query. Each candidate $y_i$ comprises observations from a nonempty subset $\mathcal{O}_i \subseteq \mathcal{M}$ of the supported modalities. The modality set $\mathcal{M}$ may include visual, audio, subtitle, or other available streams, and $M=|\mathcal{M}|$ denotes the number of supported modalities. Subtitles are candidate-side observations and are therefore distinct from the caption used as the retrieval query.

A query encoder $f_{\mathrm{q}}$ and modality-specific encoders $\{f_m\}_{m\in\mathcal{M}}$ produce $d$-dimensional representations in a common embedding space. Each encoder ends with a learned linear projection to this shared space. Although these encoders are parameterised separately, the retrieval objective aligns their outputs for cross-modal comparison. The resulting normalised embeddings are
$\mathbf{q}_i=\operatorname{norm}(f_{\mathrm{q}}(c_i))$ and $\mathbf{x}_i^m=\operatorname{norm}(f_m(y_i^m))$.
Here, $\operatorname{norm}(\mathbf{x})=\mathbf{x}/\max(\|\mathbf{x}\|_2,\epsilon_{\mathrm{n}})$,
where $\epsilon_{\mathrm{n}}>0$ is a numerical-stability constant. When $\|\mathbf{x}\|_2>\epsilon_{\mathrm{n}}$, this operation produces a unit-norm embedding, and inner products between the resulting embeddings correspond to cosine similarities.

Using these embeddings, we represent each observed candidate--modality pair as a vertex: $ \mathcal{V}=\{(i,m):1\leq i\leq N,\ m\in\mathcal{O}_i\}$. Accordingly, the number of vertices is $ n_{\mathrm{v}}=\sum_{i=1}^{N}|\mathcal{O}_i|$.
This construction preserves modality identity while allowing different observations of the same candidate to participate in distinct interactions. Because vertices are instantiated only for observed modalities, missing modalities require neither imputation nor masking; they are absent from both propagation and pooling.

Query embeddings are not propagated as vertex features. Instead, stop-gradient query embeddings are used only to define semantic associations among candidates. This design allows caption semantics to guide the construction of the training-time hypergraph without injecting query information directly into candidate representations. Consequently, each candidate can be encoded independently of the test query at inference.

\subsection{Semantic Connectivity and Learned Weights}
\label{sec:connectivity}

Cross-candidate interactions are restricted to semantically related training instances. Let $\mathbf{t}_i=\operatorname{sg}(\mathbf{q}_i)$ denote the stop-gradient caption embedding, where $\operatorname{sg}$ is the stop-gradient operator. For candidate $i$, let $\mathcal{K}(i)$ contain the indices of the $k_{\mathrm{eff}}$ other candidates with the most similar caption embeddings under the inner product $\mathbf{t}_i^\top\mathbf{t}_j$. We retain an association only when this neighbourhood relation is reciprocal:
\begin{equation}
S_{ij} = \mathbb{1}[j\in\mathcal{K}(i)] \mathbb{1}[i\in\mathcal{K}(j)] F_{ij}D_{ij}, \qquad S_{ii}=0.
\label{eq:connectivity}
\end{equation}

Here, $\mathbb{1}[\cdot]$ is the indicator function, $F_{ij}=F_{ji}$ is an optional similarity filter, and $D_{ij}=D_{ji}$ is a symmetric training-time dropout mask. In the absence of filtering or dropout, the corresponding factors are set to one. The reciprocity requirement eliminates one-sided associations supported by only one neighbourhood ranking.

{\emergencystretch=1em
In the implementation, $k_{\mathrm{eff}}=\allowbreak\min(k,\allowbreak\max(2,\lfloor N/4\rfloor))$, where $k$ is the requested neighbourhood size, and semantic connections are omitted when $N\leq k_{\mathrm{eff}}+1$. The optional filter retains similarities above
$\mu+\eta\sigma$, where $\mu$ and $\sigma$ are the mean and standard deviation of the off-diagonal caption similarities
and $\eta>0$ controls selectivity. These choices determine the candidate associations available to all relation families.
\par}

Connectivity and contribution strength are treated separately. A learned projection determines the weight of each retained
candidate pair, while a modality-dependent predictor estimates the contribution of an individual vertex:
\begin{equation}
\begin{aligned}
\mathbf{a}_i &= \operatorname{norm}(W_{\mathrm{a}}\mathbf{t}_i), & \omega_{ij}
&= S_{ij} \exp\left( \frac{\mathbf{a}_i^\top\mathbf{a}_j-1}{\tau_{\mathrm{a}}} \right),
\\
\tau_{\mathrm{a}} &=\exp(\xi),
&
\beta_i^m
&= 2\operatorname{sigmoid} \left( \mathbf{w}_{\mathrm{v}}^\top\mathbf{x}_i^m+b_{\mathrm{v}} \right).
\end{aligned}
\label{eq:weights}
\end{equation}
The parameters are
$W_{\mathrm{a}}\in\mathbb{R}^{d\times d}$,
$\xi\in\mathbb{R}$,
$\mathbf{w}_{\mathrm{v}}\in\mathbb{R}^{d}$, and
$b_{\mathrm{v}}\in\mathbb{R}$.
The connection weights satisfy $\omega_{ij}=\omega_{ji}$ and lie in $[0,1]$ in exact arithmetic, while $\beta_i^m\in(0,2)$.
Unlike a neighbourhood softmax, the exponential kernel does not require the outgoing weights of a candidate to sum to one.
It therefore preserves an absolute measure of affinity within the learned caption space.

We initialise $W_{\mathrm{a}}$ to the identity, $\tau_{\mathrm{a}}$ to $0.5$, and $\mathbf{w}_{\mathrm{v}}$ and $b_{\mathrm{v}}$ to zero. Consequently, all observed vertices initially have unit weight. The weighting parameters receive gradients from the training objective, whereas the detached caption inputs do not transmit these gradients to the query encoder. Vertex weights
remain differentiable with respect to their candidate features.

\subsection{A Common Hypergraph Propagation Formulation}
\label{sec:framework}

A hyperedge is represented by an indexed interaction $e=(\mathcal{V}_e,r_e,\omega_e,\pi_e)$. Its participating vertices form $\mathcal{V}_e\subseteq\mathcal{V}$, $r_e$ identifies its relation type, $\omega_e$ is a scalar interaction weight,
and $\pi_e(v)$ assigns a participant role to vertex $v$. Different indexed interactions may share the same vertex set while carrying different relation labels or role assignments.

Let $\mathcal{E}(v)$ denote the hyperedges incident to $v$, and let $\mathbf{h}_v^{(\ell)}\in\mathbb{R}^{d}$ be its representation at layer $\ell$. We express refinement through a message function $\mathcal{M}_{\ell}$, an aggregation operator $\mathcal{A}_{\ell}$, and an update function $\mathcal{U}_{\ell}$:
\begin{equation}
\begin{aligned}
\mathbf{m}_{e\to v}^{(\ell)}
&=
\mathcal{M}_{\ell}
\left(
v,e,
\{\mathbf{h}_u^{(\ell)}\}_{u\in\mathcal{V}_e}
\right),
\\
\mathbf{m}_v^{(\ell)}
&=
\mathcal{A}_{\ell}
\left(
\{\mathbf{m}_{e\to v}^{(\ell)}\}_{e\in\mathcal{E}(v)}
\right),
\\
\mathbf{h}_v^{(\ell+1)}
&=
\mathbf{h}_v^{(\ell)}
+
\tanh(\gamma_\ell)
\mathcal{U}_{\ell}
\left(
\mathbf{h}_v^{(\ell)},\mathbf{m}_v^{(\ell)}
\right).
\end{aligned}
\label{eq:framework}
\end{equation}
The learned scalar $\gamma_\ell$ modulates the sign and strength of the residual correction. Hyperedge weights, memberships, relation labels, and participant roles enter through the message and aggregation functions.

This formulation separates the interaction structure from the mechanism that combines its features. Additive propagation constructs weighted hyperedge summaries and redistributes them to their members. Relational propagation instead constructs recipient-specific messages whose components depend jointly on the other participants. Both are instances of Eq.~\eqref{eq:framework}.

\subsection{Weighted Additive Hypergraph}
\label{sec:additive}

The additive configuration uses document and semantic neighbourhood hyperedges. A document hyperedge connects all observed modalities of one candidate. A semantic hyperedge connects the observed modalities of an anchor candidate and those of its retained semantic neighbours. With rows indexed by observed vertices, the corresponding incidence matrices are
\begin{equation}
\begin{aligned}
H^{\mathrm{doc}}_{(j,m),i}
&=
\mathbb{1}[j=i]\beta_j^m,
\\
H^{\mathrm{sem}}_{(j,m),i}
&=
\mathbb{1}[j=i]+\omega_{ji},
\\
H
&=
[H^{\mathrm{doc}}\mid H^{\mathrm{sem}}].
\end{aligned}
\label{eq:incidence}
\end{equation}
Document membership reflects learned modality importance. Semantic membership assigns unit weight to each observed anchor modality and a shared connection weight to each observed modality of a neighbour. Consequently, a neighbour with more observed modalities contributes more total membership mass; subsequent degree normalisation operates on this total mass.

The semantic block is omitted when the batch contains no retained semantic connections. Otherwise, one semantic hyperedge is included for every candidate, including anchor-only hyperedges for candidates without neighbours. This convention is relevant because repeated or overlapping  hyperedges affect the normalisation of incident messages.

Let $Z^{(0)}\in\mathbb{R}^{n_{\mathrm{v}}\times d}$ stack the encoder embeddings as rows. Define $D_{\mathrm{e}}=\operatorname{diag}(H^\top\mathbf{1})$ and $D_{\mathrm{v}}=\operatorname{diag}(H\mathbf{1})$, where $\mathbf{1}$ is an all-ones vector of the appropriate dimension. The additive layer is
\begin{equation}
\begin{aligned}
C^{(\ell)}
&=
D_{\mathrm{v}}^{-1}H\,
\operatorname{GELU}
\left(
D_{\mathrm{e}}^{-1}H^\top
Z^{(\ell)}W_{\mathrm{V}}^{(\ell)}
\right)
W_{\mathrm{E}}^{(\ell)},
\\
Z^{(\ell+1)}
&=
Z^{(\ell)}
+
\tanh(\gamma_\ell)\phi_\ell(C^{(\ell)}).
\end{aligned}
\label{eq:additive}
\end{equation}
Here, $W_{\mathrm{V}}^{(\ell)},W_{\mathrm{E}}^{(\ell)} \in\mathbb{R}^{d\times d}$ are learned projections, and $\phi_\ell$ is GELU in intermediate layers and the identity in the final layer. Degree entries are floored at $10^{-6}$ before inversion.

The first aggregation computes a weighted average of the vertices in each hyperedge. The second combines the resulting hyperedge representations at each vertex. The residual path preserves the incoming feature while allowing within-candidate and cross-candidate context to modify it. Hyperedge types affect connectivity, but this operator shares its feature transformations across types.

\subsection{Relational Hypergraph}
\label{sec:relational}

The relational configuration makes the interaction type explicit through typed hyperedge families. Within a candidate, a single document hyperedge connects all of its observed modalities and therefore already covers every pair of them, so no separate within-candidate pairwise hyperedges are formed.

Cross-candidate hyperedges are constructed for pairs with $\omega_{ij}>0$. Same-modality hyperedges connect matching modality types across the two candidates. Cross-modality hyperedges connect different modality types; for each unordered modality pair, both assignments to the two candidates are included when the corresponding observations are available. A joint-candidate hyperedge
contains all observed modalities from both candidates. Within-candidate hyperedges have weight $\omega_e=1$, and cross-candidate hyperedges have weight $\omega_e=\omega_{ij}$. Hyperedges with fewer than two observed vertices are discarded.

The role map $\pi_e$ identifies modality type. For joint-candidate hyperedges, it additionally identifies which of the two candidates supplies the observation. These two candidate positions use disjoint role indices. Both candidate orders are instantiated, ensuring that the construction does not privilege either endpoint of the undirected semantic association.

For relational propagation, the initial state is $\mathbf{h}_v^{(0)}=\sqrt{d}\,\mathbf{x}_v$. Let $\mathbf{p}_{\pi_e(u)}^{(\ell)}\in\mathbb{R}^{d}$ and $\mathbf{r}_{r_e}^{(\ell)}\in\mathbb{R}^{d}$ denote learned role and relation embeddings. The contribution of participant $u$ and the aggregated message to recipient $v$ are
\begin{equation}
\begin{aligned}
\mathbf{a}_{e,u}^{(\ell)}
&=
\beta_u
\left(
\alpha_\ell\mathbf{h}_u^{(\ell)}
+
(1-\alpha_\ell)\mathbf{p}_{\pi_e(u)}^{(\ell)}
\right),
\\
\mathbf{m}_v^{(\ell)}
&=
\sum_{e\in\mathcal{E}(v)}
\omega_e\,
\mathbf{r}_{r_e}^{(\ell)}
\odot
\underset{u\in\mathcal{V}_e\setminus\{v\}}{\bigodot}
\mathbf{a}_{e,u}^{(\ell)}.
\end{aligned}
\label{eq:relational}
\end{equation}
The symbol $\odot$ denotes elementwise multiplication. The learned scalar $\alpha_\ell$ is initialised to $0.5$ and is unconstrained, so the feature--role combination is affine rather than necessarily convex. Vertex weights $\beta_u$ are computed from the encoder outputs and reused across layers.

Excluding the recipient from the product makes each message depend on the other participants in the interaction. For two-vertex hyperedges, the product contains one factor. For larger hyperedges, it jointly combines several participants, allowing the message to depend on their coordinate-wise interactions. Relation embeddings modulate these products by interaction type, while role embeddings distinguish contributions across modalities and candidate positions.

Messages are summed over incident hyperedges without degree normalisation and are incorporated through
\begin{equation}
\begin{aligned}
\mathbf{h}_v^{(\ell+1)}
&=
\mathbf{h}_v^{(\ell)}
+
\tanh(\gamma_\ell)\,
\sigma_\ell\big(\operatorname{LN}_\ell(\mathbf{z}_v^{(\ell)})\big),
\\
\mathbf{z}_v^{(\ell)}
&=
\operatorname{Dropout}\big(
W_\ell
[\mathbf{m}_v^{(\ell)}\Vert\mathbf{h}_v^{(\ell)}]
+\mathbf{b}_\ell
\big),
\end{aligned}
\label{eq:relational-update}
\end{equation}
where $\Vert$ denotes concatenation, $W_\ell\in\mathbb{R}^{d\times 2d}$, $\mathbf{b}_\ell\in\mathbb{R}^{d}$, and $\operatorname{LN}_\ell$ is layer normalisation. The activation $\sigma_\ell$ is ReLU in intermediate layers and the identity in the final layer.

Concatenating the current feature with its aggregated message allows the update to depend on both local information and relational context. An observed vertex without incident hyperedges receives a zero message but may still undergo this local transformation. When the entire relational hypergraph is empty, the implementation bypasses refinement. We use at most two propagation layers.

\subsection{Retrieval Supervision and Inference}
\label{sec:training}

After $L$ layers, let $\mathbf{u}_i^m$ denote the final unnormalised modality representation. For additive propagation, it is the corresponding row of $Z^{(L)}$; for relational propagation, it is $\mathbf{h}_{(i,m)}^{(L)}/\sqrt{d}$. The module produces normalised modality embeddings and a pooled candidate representation:
\begin{equation}
\begin{aligned}
\bar{\mathbf{x}}_i^m
&=
\operatorname{norm}(\mathbf{u}_i^m),
\\
\tilde{\mathbf{h}}_i
&=
W_{\mathrm{pool}}
\frac{
\sum_{m\in\mathcal{O}_i}\beta_i^m\mathbf{u}_i^m
}{
\max\left(
\sum_{m\in\mathcal{O}_i}\beta_i^m,\epsilon_{\mathrm{p}}
\right)
},
\\
\mathbf{h}_i^{\mathrm{doc}}
&=
\operatorname{norm}(\tilde{\mathbf{h}}_i).
\end{aligned}
\label{eq:outputs}
\end{equation}
Here, $W_{\mathrm{pool}}\in\mathbb{R}^{d\times d}$ is learned and $\epsilon_{\mathrm{p}}=10^{-6}$. Pooling includes only observed modalities and uses the same learned vertex weights as the corresponding weighted refinement configuration.

Let $s_b$ denote the unscaled compatibility function of retrieval backbone $b$, where larger values indicate stronger compatibility. Refinement changes the embeddings supplied to this function while preserving its form:
\begin{equation}
\Phi_{ij}^{(b)}
=
\frac{
s_b\left(
\mathbf{q}_i,
\{\bar{\mathbf{x}}_j^m\}_{m\in\mathcal{O}_j}
\right)
}{
\vartheta_b
},
\qquad \vartheta_b>0.
\label{eq:score}
\end{equation}
The temperature convention of the backbone is retained; if no temperature scaling is used, $\vartheta_b=1$.

For a Euclidean Gramian-volume backbone,\\ $s_b=-\sqrt{\det(A_{ij}^{\top}A_{ij})}$, where $A_{ij}$ contains the query and observed candidate embeddings as columns. Other backbones retain their corresponding compatibility functions. Thus, volume scoring is one instantiation of the framework rather than a requirement of hypergraph refinement. With one designated positive pair per batch index, the training objective is
{\small
\begin{equation}
\begin{aligned}
\mathcal{L}_{\mathrm{ret}}^{(b)}
&=
-\frac{1}{2N}
\sum_{i=1}^{N}
\left[
\log
\frac{\exp(\Phi_{ii}^{(b)})}
{\sum_{j=1}^{N}\exp(\Phi_{ij}^{(b)})}
+
\log
\frac{\exp(\Phi_{ii}^{(b)})}
{\sum_{j=1}^{N}\exp(\Phi_{ji}^{(b)})}
\right],
\\
\mathcal{L}^{(b)}
&=
\mathcal{L}_{\mathrm{ret}}^{(b)}
+
\sum_{a\in\mathcal{A}_b}
\lambda_a\mathcal{L}_a^{(b)}.
\end{aligned}
\label{eq:objective}
\end{equation}
}
The set $\mathcal{A}_b$ indexes the auxiliary objectives retained from the selected training recipe, and $\lambda_a\geq0$ controls their contributions. Semantic connectivity does not alter the positive assignments: related candidates exchange information during refinement but remain contrastive negatives unless they are designated matches.

For the \hy recipe, the auxiliary terms are pooled caption--candidate alignment, variance regularisation, and the backbone's matching loss, with their original definitions~\cite{nag2027hyvol}. The pooled representation in Eq.~\eqref{eq:outputs} supplies the candidate feature for auxiliary alignment. These auxiliary choices are separate from the hypergraph propagation operator and
are held fixed when comparing refinement configurations under a common training recipe.

At inference, we replace refined embeddings with the encoder outputs $\mathbf{x}_j^m$ in Eq.~\eqref{eq:score}. The hypergraph, propagation layers, and auxiliary pooling head are not evaluated. Consequently, relational context influences retrieval through the encoder parameters learned during training, without test-time neighbourhood construction or interaction among test candidates.

\subsection{Special Cases and Instantiations}
\label{sec:instances}

The framework separates hyperedge structure, contribution weights, and propagation. The following choices recover \hy and define the weighted and relational configurations.

\paragraph{\hy} ~\cite{nag2027hyvol} is obtained by selecting document hyperedges and caption-guided semantic neighbourhood hyperedges, assigning unit document memberships, and using its original learned semantic attention. Specifically, detached caption features are projected and scored by an additive attention function; a softmax over retained mutual neighbours produces attention coefficients, which are subsequently symmetrised. These coefficients determine semantic memberships, together with unit anchor memberships.

We choose the message and aggregation functions in Eq.~\eqref{eq:framework} as \hy's vertex-to-hyperedge and hyperedge-to-vertex additive operators, with its original degree conventions and gated residual update. Relation labels and participant roles do not explicitly parameterize these messages. Selecting the original neighborhood dropout, auxiliary objectives, and volume-based scorer completes the \hy special case.

\paragraph{Weighted hypergraph (\textsc{\why}):} The weighted configuration uses the neighbourhood hyperedges in Eq.~\eqref{eq:incidence} and additive propagation in Eq.~\eqref{eq:additive}. Learned vertex weights regulate within-candidate contributions, while the symmetric exponential kernel in Eq.~\eqref{eq:weights} regulates cross-candidate connections. Unlike \hy's original attention rule, these connection weights are not obtained through a neighborhood softmax.

\paragraph{Relational hypergraph (\textsc{wRHG}):} The relational configuration uses four typed hyperedge families. Within a single candidate, the document hyperedge connects all observed modalities of that candidate; since it already contains every pair of these modalities (e.g., video with audio), separate modality-pair hyperedges are not required. Between two related candidates $i$ and $j$, i.e., candidates whose captions are mutual top-$k$ neighbours, three families are used: same-modality hyperedges connect matching modalities (e.g., video of $i$ with video of $j$), cross-modality hyperedges connect different modalities (e.g., video of $i$ with audio of $j$), and the joint-candidate hyperedge contains all observed modalities of both candidates. Messages over these hyperedges are formed multiplicatively by Eq.~\eqref{eq:relational}, and vertex features are updated by Eq.~\eqref{eq:relational-update}. Each family has its own relation embedding, and each participating modality has its own role embedding, so the model distinguishes, for example, a video--video same-modality interaction from a video--audio cross-modality interaction, which additive propagation would combine through the same transformation.

These configurations are instances of the common propagation formulation rather than successive restrictions of a single multiplicative operator. \hy is recovered by selecting its complete structural and computational specification. Restricting relation families alone changes the available interactions but does not convert multiplicative propagation into \hy's additive aggregation.

\section{Experiments}
\label{sec:Experiments}
We evaluate whether training-time hypergraph fusion improves multimodal retrieval across different backbones, benchmarks, and modality configurations. Our experiments assess zero-shot retrieval performance and robustness to missing modalities, comparing \hy, \why, and \wrhg with their corresponding backbones and existing retrieval methods. We first describe the datasets and experimental protocol, then present and analyze the results.
\subsection{Datasets}
\label{sec:data}
We describe the training corpus and evaluation benchmarks used to assess retrieval performance across different modality configurations.
\subsubsection{Training data}
\label{sec:data_train}

All models trained in this work are initialized from the \vast checkpoint~\cite{chen2023vast} and trained on the $150$K-clip subset of \vast-$27$M used by \gram~\cite{cicchetti2025gramian} and \hypergram~\cite{na2026hypergram}. Each clip has five visual captions and audio captions, a subtitle containing the speech transcript, and one omni-modal caption that integrates visual, acoustic, and speech information. We use the omni-modal caption as the training text to provide supervision that captures complementary information across the candidate modalities. The annotation file for the subset lists $150,154$ clips, of which $136,694$ were successfully recovered with both video and audio because online video availability changed. These recovered clips constitute the effective training corpus for all models trained in this work.

For \gram~\cite{cicchetti2025gramian}, \tri~\cite{cicchetti2026triangle}, and PMRL~\cite{liu2026principled}, we evaluated the checkpoints released by their authors. The exact set of clips available during the original training of these checkpoints cannot be reconstructed. We therefore distinguish comparisons against released checkpoints from comparisons between models trained on our common accessible subset.

\subsubsection{Evaluation benchmarks}
\label{sec:data_eval}

Table~\ref{tab:datasets} summarises the eleven evaluation benchmarks: eight for video--text retrieval and three for audio--text retrieval. These benchmarks differ in their temporal structure, annotation granularity, and available modalities, enabling evaluation across short videos, multi-shot narratives, long-video excerpts, and sound recordings. We do not use any target benchmarks for training in this work, and we perform all evaluations without target-specific adaptation. When several alternative captions describe an item, we select the first caption in the annotation file. DiDeMo and ActivityNet instead use paragraph queries constructed from the sentences associated with each video, as detailed below. All methods we evaluate use the same queries and candidate collections.

\paragraph{Video--text retrieval.}
MSR-VTT~\cite{MSRVTT} contains short videos paired with natural-language descriptions. We evaluate on the 1,000-video test split used by \vast~\cite{chen2023vast} and \gram~\cite{cicchetti2025gramian}. We evaluate DiDeMo~\cite{DIDEMO} and ActivityNet~\cite{ACTIVITYNET} as paragraph-to-video retrieval tasks by concatenating the annotated sentences associated with each video into a single query. These settings assess retrieval using descriptions that can encompass multiple events within a video. Following \vast and \gram, we evaluate on 1,003 videos for DiDeMo and 4,917 videos for ActivityNet.

VATEX~\cite{wang2019vatex} supports evaluation with visual, audio, and subtitle information. We follow the test split adopted by \vast and \gram; however, when we collected the data, only 431 videos from this split remained accessible. All methods we evaluate use this identical subset. We therefore treat published results from a different video collection as contextual references rather than directly matched comparisons.
\begin{table*}[t]
\centering
\caption{Statistics of the evaluation benchmarks. $T$, $V$, $A$, and $S$ denote text, video, audio, and subtitles. \emph{Mode} lists the modalities compared with the text, with the primary modality first: T--V, T--VA, and T--VAS on the video benchmarks, and T--A and T--AV on the audio benchmarks. Retrieval is evaluated in both directions: text-to-video and video-to-text on the video benchmarks, and text-to-audio and audio-to-text on the audio benchmarks. \emph{Text query} is the annotation used on the text side. Train and Val give the split sizes; we do not use them, since all evaluations are zero-shot. Test is the number of evaluated items. \emph{Frames} is the number of video frames and \emph{Audio} the number of 10\,s audio segments sampled per clip. \emph{Max caption length} is the maximum number of text tokens kept per query; longer queries are truncated.}
\label{tab:datasets}
\footnotesize
\setlength{\tabcolsep}{4pt}
\resizebox{0.9\textwidth}{!}{%
\begin{tabular}{@{}cccccccccc@{}}
\toprule
Dataset & Streams & Mode & Text query & Train & Val & Test & Frames & \begin{tabular}[c]{@{}c@{}}Audio\\(10\,s)\end{tabular} & \begin{tabular}[c]{@{}c@{}}Max caption\\length\end{tabular} \\
\midrule
\multicolumn{10}{c}{\emph{Video--text retrieval}} \\
\midrule
MSR-VTT~\cite{MSRVTT}                   & $V,A,S$ & T--V, T--VA, T--VAS & caption                  & 9{,}000  & --      & 1{,}000 & 8  & 1  & 40 \\
DiDeMo~\cite{DIDEMO}                    & $V,A$   & T--V, T--VA         & paragraph                & 8{,}394  & 1{,}065 & 1{,}003 & 8  & 2  & 70 \\
ActivityNet~\cite{ACTIVITYNET}          & $V,A$   & T--V, T--VA         & paragraph                & 10{,}009 & --      & 4{,}917 & 8  & 2  & 70 \\
VATEX~\cite{wang2019vatex}              & $V,A,S$ & T--V, T--VA, T--VAS & caption                  & 14{,}060 & --      & 431     & 8  & 1  & 40 \\
Shot2Story~\cite{han2023shot2story}     & $V,A$   & T--V, T--VA         & video summary            & 14{,}016 & 1{,}982 & 4{,}023 & 8  & 1  & 70 \\
MSVD~\cite{chen2011msvd}                & $V$     & T--V                & caption                  & 1{,}200  & 100     & 670     & 8  & -- & 40 \\
FLARE~\cite{flare2026}                  & $V,A$   & T--V, T--VA         & vision / unified caption & --       & --      & 5{,}000 & 8  & 1  & 70 \\
VALOR-32K~\cite{liu2024valor}           & $V,A$   & T--V, T--VA         & audio-visual caption     & 25{,}327 & 3{,}500 & 3{,}491 & 8  & 1  & 70 \\
\midrule
\multicolumn{10}{c}{\emph{Audio--text retrieval}} \\
\midrule
AudioCaps~\cite{AUDIOCAPS}              & $V,A$   & T--AV               & audio caption            & --       & --      & 700     & 8  & 1  & 40 \\
Clotho~\cite{drossos2020clotho}         & $A$     & T--A                & caption                  & 3{,}839  & 1{,}045 & 1{,}045 & -- & 1  & 40 \\
SoundDescs~\cite{koepke2023audio}       & $A$     & T--A                & description              & 23{,}085 & 4{,}947 & 4{,}947 & -- & 2  & 70 \\
\bottomrule
\end{tabular}}
\end{table*}

Shot2Story-20K~\cite{han2023shot2story} contains multi-shot videos paired with summary descriptions. Its 4,023-video test split allows us to examine retrieval when a query describes content distributed across several shots. MSVD~\cite{chen2011msvd} contains short videos with multiple textual descriptions, and we evaluate its 670-video test split. Since audio is unavailable in our MSVD evaluation data, only the T--V setting is reported.

FLARE~\cite{flare2026} provides clips extracted from longer videos, together with vision, audio, and unified audiovisual captions. We evaluate a fixed subset of 5,000 clips, using vision captions for T--V and unified captions for T--VA. These settings differ in both query content and candidate modalities; therefore, their performance differences do not isolate the contribution of audio.

VALOR-32K~\cite{liu2024valor} consists of approximately 10\,s clips accompanied by captions describing both visible content and audible events. We evaluate its 3,491-item test split. Its audiovisual annotations complement benchmarks whose descriptions primarily emphasize visual content.

\paragraph{Audio--text retrieval.}
AudioCaps~\cite{AUDIOCAPS} pairs audio events with natural-language descriptions. We use the same 700-item test split as \vast~\cite{chen2023vast} and \gram~\cite{cicchetti2025gramian}. The corresponding video is available alongside the audio, allowing evaluation in the T--AV setting.

Clotho v2.1~\cite{drossos2020clotho} contains audio recordings with five captions per recording. We evaluate its 1,045-item evaluation split using the first caption associated with each recording. SoundDescs~\cite{koepke2023audio} contains sound-effect recordings paired with textual descriptions, and we evaluate on 4,947 test items. Both Clotho and SoundDescs are evaluated in the T--A setting, assessing retrieval from acoustic information alone.

\subsection{Compared Methods and Implementation Details}
\label{sec:implementation}

\subsubsection{Compared methods}
\label{sec:compared_methods}

We evaluate the proposed refinement framework with two retrieval backbones, \gram \footnote{\url{https://github.com/ispamm/GRAM}}
~\cite{cicchetti2025gramian} and \hypergram \footnote{\url{https://github.com/uta-smile/hypergram}}~\cite{na2026hypergram}. \gram measures multimodal compatibility through Gramian volumes, whereas \hypergram combines Euclidean and hyperbolic volume-based representations. Evaluating both backbones allows us to examine whether the benefits of hypergraph refinement persist across these two geometric scoring formulations. In each case, the refinement module operates on the candidate representations during training while retaining the backbone's retrieval scoring function.

For each backbone, we compare the model without hypergraph refinement against three configurations: \hy \footnote{\url{https://github.com/CVML-CFU/HyVol}} \cite{nag2027hyvol}, the weighted hypergraph configuration \why, and the weighted relational configuration \wrhg. \hy provides the reference configuration with within-candidate and caption-guided semantic hyperedges. The \why configuration introduces learned vertex contributions and symmetric semantic affinity weights within additive propagation. The \wrhg configuration additionally distinguishes relation types and participant roles through relational propagation. These comparisons evaluate the full configurations; Section~\ref{sec:neighborhood} also examines the hypergraph's sensitivity to neighborhood size $k$.

We additionally compare against \tri\footnote{\url{https://github.com/ispamm/TRIANGLE/}}~\cite{cicchetti2026triangle} and \pmrl\footnote{\url{https://github.com/Xiaohao-Liu/PMRL}}~\cite{liu2026principled}. \tri offers an alternative volume-based retrieval approach using triangular area, while \pmrl provides an additional multimodal representation-learning baseline. We include results from other published retrieval systems to contextualize performance within the broader literature.

\subsubsection{Comparison protocol}
\label{sec:comparison_protocol}

We distinguish between methods evaluated using released checkpoints, methods trained in this work, and results taken from prior publications. For \gram, \tri, and \pmrl, we use the checkpoints released by their authors. Since a pretrained \hypergram checkpoint is unavailable, we train \hypergram using the authors' public implementation and training recipe. All models trained in this work start from the \vast initialization and use the effective training corpus described in Section~\ref{sec:data_train}.

All checkpoints evaluated by us are tested on the same benchmark items and query annotations within each retrieval setting. We evaluate \gram using its retrieval protocol, and we use \tri's released evaluation code. Any method-specific scoring or post-processing is retained and must therefore be considered when interpreting comparisons across methods.

The common evaluation data remove differences arising from query selection and candidate availability. However, the released checkpoints may differ from models trained in this work in their effective training samples and optimization history. Comparisons against these checkpoints therefore measure performance relative to the available reference systems, rather than isolating the effect of a single architectural component.

We report published scores separately as reference results because their evaluation collections, input processing, or training conditions may differ from ours. This distinction is particularly relevant for benchmarks affected by video availability, including the reduced VATEX test subset.

\subsubsection{Input processing and model configuration}
\label{sec:model_configuration}

\paragraph{Backbone encoders.} All compared models share the modality encoders of \vast~\cite{chen2023vast}: EVA-CLIP ViT-g/14\footnote{\url{https://huggingface.co/QuanSun/EVA-CLIP}} for video frames, BEATs\footnote{\url{https://github.com/microsoft/unilm/tree/master/beats}} for audio, and BERT-base\footnote{\url{https://huggingface.co/google-bert/bert-base-uncased}} for both captions and subtitles. We remove the \vast fusion layers, project each stream to a $512$-dimensional space, and $\ell_2$-normalize it so that the backbone score acts directly on the per-modality embeddings. The backbone temperature is loaded from its checkpoint and kept trainable.

\paragraph{Training recipe.} \textsc{\hy}, \textsc{\why} and \textsc{\wrhg} share the same optimization and data settings on both backbones. Starting from the \vast checkpoint, we train each model for one epoch (533 steps) on the recovered training corpus with a global batch of 256 clips, sampling two frames and one 10\,s audio segment per clip. No encoder is frozen. We use AdamW with $\beta=(0.9,0.98)$, weight decay $0.01$ and gradient-norm clipping at $2.0$. The learning rate is $2\times10^{-5}$ for the backbone, $5\times10^{-7}$ for the EVA-CLIP visual encoder and $5\times10^{-4}$ for the hypergraph module, with linear warm-up over the first $10\%$ of the steps followed by linear decay.

\paragraph{Hypergraph and loss settings.} We split the batch across four GPUs, and each GPU builds its hypergraph from its own 64 clips, so each GPU's batch also forms the pool of semantic neighbors. All three configurations use $k=12$, which gives $k_{\mathrm{eff}}=12$ in Eq.~\eqref{eq:connectivity}, no similarity filter, a semantic-edge dropout of $0.3$, and $L=2$ propagation layers. The auxiliary weights in Eq.~\eqref{eq:objective} are $\lambda_{\mathrm{doc}}=1.0$ for pooled caption--candidate alignment, $\lambda_{\mathrm{reg}}=0.1$ for variance regularisation (with stability constant $10^{-4}$), and $\lambda_{\mathrm{itm}}=0.1$ for the matching loss. The volume-based retrieval and alignment losses use label smoothing of $0.1$. Table~\ref{tab:hyper_parameters} in Appendix~\ref{app:hparams} lists all hyperparameters.

\paragraph{Differences across configurations.} The three configurations differ only in their hypergraph weighting and propagation, as defined in Section~\ref{sec:instances}. For all three configurations on \textsc{\hypergram}, we retain the authors' hybrid-geometry settings: a learnable curvature initialized to $1$, learnable Euclidean and hyperbolic weights initialized to $0.5$, and hyperbolic gradient clipping at $1.0$.

\paragraph{Hypergraph at training and inference.} We apply hypergraph refinement to the observed candidate modalities during training. Training captions guide the construction of semantic neighborhoods but are not included as candidate vertices. Missing modalities are excluded from the hypergraph, so they contribute neither vertex features nor propagated messages. At inference, we discard the hypergraph, propagation layers, and auxiliary pooling head, and retrieve scores using the corresponding backbone's scoring function in Eq.~\eqref{eq:score}. Consequently, the refinement configurations introduce no additional hypergraph computation during retrieval and require no neighborhood construction over evaluation candidates.

\paragraph{Input processing.} Table~\ref{tab:datasets} lists the sampling used at test time. Eight frames are sampled for every candidate that contains video, and audio is represented by one or two 10\,s segments depending on the benchmark. Queries are truncated to 40 or 70 tokens, which accommodates the different clip durations and description lengths of the benchmarks.

\paragraph{Checkpoint selection.} We validate ten times during the epoch on the MSR-VTT 1k split in the T--VAS setting and keep the checkpoint with the highest text-to-video R@1 of the matching score. MSR-VTT is therefore the only benchmark used for model selection; the other ten benchmarks are used exclusively for testing. The neighbourhood size $k$ was likewise chosen from the sweep in Section~\ref{sec:neighborhood}, which uses four of the evaluation benchmarks; a single value is used for all benchmarks and configurations.

\paragraph{Hardware and runs.} All models are trained on four NVIDIA A100 GPUs with fp16 mixed precision, while hypergraph aggregation is computed in float32 for numerical stability. Following \gram~\cite{cicchetti2025gramian} and \hypergram~\cite{na2026hypergram},  which report results from a single training run, we train each configuration once.

\subsection{Zero-Shot Retrieval Performance}
\label{sec:main_results_1}

We evaluate retrieval performance across the eleven benchmarks described in Section~\ref{sec:data_eval}. Tables~\ref{tab:t2v_wacv} and~\ref{tab:v2t_wacv} report text-to-video and video-to-text retrieval, respectively, while Table~\ref{tab:audio_ret} presents both directions of audio--text retrieval. We first examine improvements over the corresponding backbone, then compare refinement configurations and their behavior across modalities. All improvements discussed below are absolute differences in R@1, expressed in percentage points.

\subsubsection{Video--text retrieval}
\label{sec:video_result_2}

\begin{figure*}[t]
\centering
\includegraphics[width=0.85\textwidth]{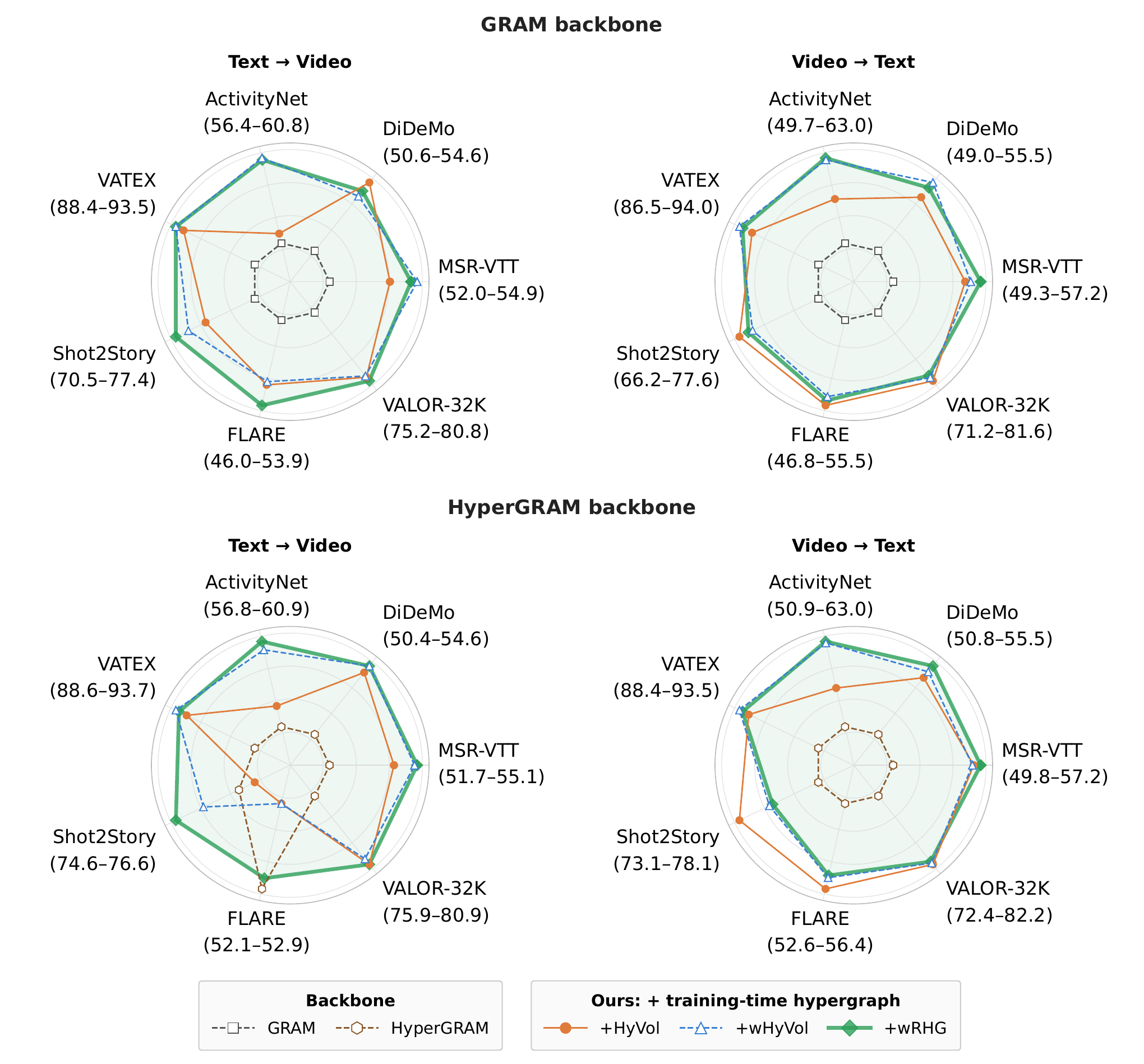}
\caption{Zero-shot text-to-video and video-to-text retrieval (R@1) with the \textsc{\gram} backbone (top) and the \textsc{\hypergram} backbone (bottom) on seven video benchmarks, all evaluated in the T--VA setting. Each axis is scaled independently to the range of R@1 values on that benchmark; the minimum and maximum R@1 are shown in parentheses below its name, so distances are comparable only within an axis. The exact values are the T--VA rows of Tables~\ref{tab:t2v_wacv} and~\ref{tab:v2t_wacv}.}

\label{fig:tva_radar}
\end{figure*}

\paragraph{Performance across backbones and retrieval directions.}
Hypergraph refinement improves retrieval across a broad range of benchmarks. With \gram, \wrhg outperforms the backbone in all 17 text-to-video settings and all 17 video-to-text settings. With \hypergram, it improves 16 of the 17 text-to-video settings and all 17 video-to-text settings. These results show that the benefits extend across both backbone formulations and retrieval directions, although their magnitude depends on the benchmark and available modalities. Figure~\ref{fig:tva_radar} visualizes these results in the T--VA setting, with the minimum and maximum R@1 of each benchmark shown below its name.

ActivityNet exhibits particularly substantial improvements. In the T--VA setting, \gram+\wrhg increases text-to-video R@1 from 56.4 to 60.7 and video-to-text R@1 from $49.7$ to $63.0$. \hypergram + \wrhg shows a similar pattern, increasing the corresponding scores from $56.8$ to $60.9$ and from $50.9$ to $63.0$. The larger gains in video-to-text retrieval indicate that the effect of refinement is not symmetric across retrieval directions. Although both directions use the same compatibility scores, ranking candidates for each textual query and ranking descriptions for each video impose different retrieval conditions.

\paragraph{Effect of weighting and relational modeling.} The weighted configurations improve on \hy~\cite{nag2027hyvol} in several settings. On ActivityNet T--VA, replacing \hy with \why increases \gram text-to-video R@1 from 56.9 to 60.8 and video-to-text R@1 from 56.6 to 62.7. With \hypergram, the corresponding scores increase from 57.8 to 60.5 and from 56.4 to 62.8. The consistent pattern across backbones indicates that the weighted configuration is particularly effective on ActivityNet.

Relational modeling provides further benefits in selected settings. On Shot2Story text-to-video retrieval with \gram, \wrhg reaches 76.7 in T--V and 77.4 in T--VA, compared with 75.6 and 76.3 for \why. On MSR-VTT video-to-text retrieval in T--V, \wrhg improves over \why from 55.1 to 56.5 with \gram and from 55.2 to 55.9 with \hypergram. These comparisons identify settings in which explicit modeling of distinct interactions provides improvements over weighted additive propagation.

The ordering of the configurations is nevertheless not uniform. \hy~\cite{nag2027hyvol} achieves the highest text-to-video score on MSVD for both backbones, reaching 75.8. It also outperforms the weighted configurations on Shot2Story video-to-text retrieval and VALOR-32K video-to-text retrieval. Thus, the additional weighting and relational mechanisms provide benchmark-dependent benefits rather than consistent improvement across settings. Section~\ref{sec:neighborhood} examines one such design choice, the neighbourhood size, under otherwise fixed settings.

\paragraph{Influence of available modalities.}
Additional candidate streams can improve retrieval when they provide information relevant to the query. On MSR-VTT, \gram + \wrhg increases text-to-video R@1 from $53.0$ with video alone to $54.7$ with video and audio, and to $57.4$ when subtitles are also included. \hypergram + \wrhg follows the same trend, achieving $54.2$, $55.1$, and $57.3$, respectively.

The contribution of audio is particularly pronounced on VALOR-32K, whose descriptions cover both visible and audible events. Adding audio increases text-to-video R@1 from $70.2$ to $80.8$ for \gram + \wrhg and from $70.4$ to $80.9$ for \hypergram + \wrhg. This pattern is consistent with the audiovisual content of the queries and illustrates the importance of evaluating fusion under annotations that describe the available streams.

Additional modalities do not guarantee higher performance. On FLARE, \hypergram+\wrhg obtains $52.8$ in T--VA, slightly below the backbone score of $52.9$. Moreover, FLARE uses different descriptions for T--V and T--VA, so we cannot attribute their difference solely to the inclusion of audio. These results highlight the dependence of retrieval performance on the combination of candidate information, query content, and backbone.

\begin{table*}[t]
\centering
\caption{Zero-shot text-to-video retrieval (R@1) on eight video benchmarks. For FLARE, T--V uses the vision captions and T--VA the unified captions. The compared models \textsc{Triangle}~\cite{cicchetti2026triangle}, \textsc{Gram}~\cite{cicchetti2025gramian} and \textsc{HyperGram}~\cite{na2026hypergram} are all volume-based: they score a clip by the area of a triangle, the Gramian volume, and a combination of Euclidean and hyperbolic volumes, respectively; +\textsc{HyVol}, +\textsc{wHyVol} and +\textsc{wRHG} denote the backbone trained with HyVol~\cite{nag2027hyvol}, weighted HyVol and the weighted relational hypergraph, respectively. $^\star$Authors' released checkpoint (\textsc{Triangle}, \textsc{Gram}), evaluated by us on the same test data: \textsc{Gram} with its retrieval protocol, \textsc{Triangle} with its released evaluation code. $^\dag$Trained by us with the authors' released code and recipe (\textsc{HyperGram}). Numbers in parentheses give the change with respect to the backbone (\textsc{Gram} or \textsc{HyperGram}); green marks an improvement, red a drop. A dash (--) marks a setting that is not available. Bold marks the best result within each backbone group.}
\label{tab:t2v_wacv}\label{tab:t2v_new}
\footnotesize
\setlength{\tabcolsep}{3pt}
\resizebox{0.9\textwidth}{!}{%
\begin{tabular}{@{}llccccccccc@{}}
\toprule
 & & Baseline & \multicolumn{4}{c}{\textsc{Gram} backbone} & \multicolumn{4}{c}{\textsc{HyperGram} backbone} \\
\cmidrule(lr){3-3}\cmidrule(lr){4-7}\cmidrule(lr){8-11}
Dataset & Mode & \textsc{Triangle}$^\star$ & \textsc{Gram}$^\star$ & +\textsc{HyVol} & +\textsc{wHyVol} & +\textsc{wRHG} & \textsc{HyperGram}$^\dag$ & +\textsc{HyVol} & +\textsc{wHyVol} & +\textsc{wRHG}  \\
\midrule
MSR-VTT~\cite{MSRVTT} & T--V & -- & 51.9 & 52.6\,{\scriptsize\textcolor{green!50!black}{(+0.7)}} & \textbf{53.4}\,{\scriptsize\textcolor{green!50!black}{(+1.5)}} & 53.0\,{\scriptsize\textcolor{green!50!black}{(+1.1)}} & 50.8 & 52.5\,{\scriptsize\textcolor{green!50!black}{(+1.7)}} & \textbf{54.3}\,{\scriptsize\textcolor{green!50!black}{(+3.5)}} & 54.2\,{\scriptsize\textcolor{green!50!black}{(+3.4)}}  \\
 & T--VA & 50.5 & 52.0 & 54.0\,{\scriptsize\textcolor{green!50!black}{(+2.0)}} & \textbf{54.9}\,{\scriptsize\textcolor{green!50!black}{(+2.9)}} & 54.7\,{\scriptsize\textcolor{green!50!black}{(+2.7)}} & 51.7 & 54.2\,{\scriptsize\textcolor{green!50!black}{(+2.5)}} & 55.0\,{\scriptsize\textcolor{green!50!black}{(+3.3)}} & \textbf{55.1}\,{\scriptsize\textcolor{green!50!black}{(+3.4)}}  \\
 & T--VAS & -- & 51.8 & 55.8\,{\scriptsize\textcolor{green!50!black}{(+4.0)}} & 56.8\,{\scriptsize\textcolor{green!50!black}{(+5.0)}} & \textbf{57.4}\,{\scriptsize\textcolor{green!50!black}{(+5.6)}} & 52.3 & 56.1\,{\scriptsize\textcolor{green!50!black}{(+3.8)}} & 57.0\,{\scriptsize\textcolor{green!50!black}{(+4.7)}} & \textbf{57.3}\,{\scriptsize\textcolor{green!50!black}{(+5.0)}}  \\
\midrule
DiDeMo~\cite{DIDEMO} & T--V & -- & 51.3 & 53.8\,{\scriptsize\textcolor{green!50!black}{(+2.5)}} & \textbf{54.1}\,{\scriptsize\textcolor{green!50!black}{(+2.8)}} & 54.0\,{\scriptsize\textcolor{green!50!black}{(+2.7)}} & 50.0 & 53.6\,{\scriptsize\textcolor{green!50!black}{(+3.6)}} & \textbf{54.3}\,{\scriptsize\textcolor{green!50!black}{(+4.3)}} & 54.0\,{\scriptsize\textcolor{green!50!black}{(+4.0)}}  \\
 & T--VA & 51.0 & 50.6 & \textbf{54.6}\,{\scriptsize\textcolor{green!50!black}{(+4.0)}} & 53.8\,{\scriptsize\textcolor{green!50!black}{(+3.2)}} & 54.1\,{\scriptsize\textcolor{green!50!black}{(+3.5)}} & 50.4 & 54.2\,{\scriptsize\textcolor{green!50!black}{(+3.8)}} & \textbf{54.6}\,{\scriptsize\textcolor{green!50!black}{(+4.2)}} & \textbf{54.6}\,{\scriptsize\textcolor{green!50!black}{(+4.2)}}  \\
\midrule
ActivityNet~\cite{ACTIVITYNET} & T--V & -- & 55.5 & 56.2\,{\scriptsize\textcolor{green!50!black}{(+0.7)}} & \textbf{59.6}\,{\scriptsize\textcolor{green!50!black}{(+4.1)}} & \textbf{59.6}\,{\scriptsize\textcolor{green!50!black}{(+4.1)}} & 56.1 & 56.7\,{\scriptsize\textcolor{green!50!black}{(+0.6)}} & 59.5\,{\scriptsize\textcolor{green!50!black}{(+3.4)}} & \textbf{59.9}\,{\scriptsize\textcolor{green!50!black}{(+3.8)}}  \\
 & T--VA & 56.0 & 56.4 & 56.9\,{\scriptsize\textcolor{green!50!black}{(+0.5)}} & \textbf{60.8}\,{\scriptsize\textcolor{green!50!black}{(+4.4)}} & 60.7\,{\scriptsize\textcolor{green!50!black}{(+4.3)}} & 56.8 & 57.8\,{\scriptsize\textcolor{green!50!black}{(+1.0)}} & 60.5\,{\scriptsize\textcolor{green!50!black}{(+3.7)}} & \textbf{60.9}\,{\scriptsize\textcolor{green!50!black}{(+4.1)}}  \\
\midrule
VATEX~\cite{wang2019vatex} & T--V & -- & 87.2 & 92.1\,{\scriptsize\textcolor{green!50!black}{(+4.9)}} & 91.9\,{\scriptsize\textcolor{green!50!black}{(+4.7)}} & \textbf{92.3}\,{\scriptsize\textcolor{green!50!black}{(+5.1)}} & 88.4 & \textbf{92.6}\,{\scriptsize\textcolor{green!50!black}{(+4.2)}} & 92.3\,{\scriptsize\textcolor{green!50!black}{(+3.9)}} & 92.3\,{\scriptsize\textcolor{green!50!black}{(+3.9)}}  \\
 & T--VA & 88.9 & 88.4 & 93.0\,{\scriptsize\textcolor{green!50!black}{(+4.6)}} & \textbf{93.5}\,{\scriptsize\textcolor{green!50!black}{(+5.1)}} & \textbf{93.5}\,{\scriptsize\textcolor{green!50!black}{(+5.1)}} & 88.6 & 93.0\,{\scriptsize\textcolor{green!50!black}{(+4.4)}} & \textbf{93.7}\,{\scriptsize\textcolor{green!50!black}{(+5.1)}} & 93.5\,{\scriptsize\textcolor{green!50!black}{(+4.9)}}  \\
 & T--VAS & -- & 88.4 & \textbf{94.0}\,{\scriptsize\textcolor{green!50!black}{(+5.6)}} & 93.5\,{\scriptsize\textcolor{green!50!black}{(+5.1)}} & \textbf{94.0}\,{\scriptsize\textcolor{green!50!black}{(+5.6)}} & 89.8 & 93.3\,{\scriptsize\textcolor{green!50!black}{(+3.5)}} & \textbf{93.5}\,{\scriptsize\textcolor{green!50!black}{(+3.7)}} & \textbf{93.5}\,{\scriptsize\textcolor{green!50!black}{(+3.7)}}  \\
\midrule
Shot2Story~\cite{han2023shot2story} & T--V & -- & 70.5 & 75.1\,{\scriptsize\textcolor{green!50!black}{(+4.6)}} & 75.6\,{\scriptsize\textcolor{green!50!black}{(+5.1)}} & \textbf{76.7}\,{\scriptsize\textcolor{green!50!black}{(+6.2)}} & 75.5 & 76.0\,{\scriptsize\textcolor{green!50!black}{(+0.5)}} & 76.3\,{\scriptsize\textcolor{green!50!black}{(+0.8)}} & \textbf{76.5}\,{\scriptsize\textcolor{green!50!black}{(+1.0)}}  \\
 & T--VA & 75.8 & 70.5 & 74.8\,{\scriptsize\textcolor{green!50!black}{(+4.3)}} & 76.3\,{\scriptsize\textcolor{green!50!black}{(+5.8)}} & \textbf{77.4}\,{\scriptsize\textcolor{green!50!black}{(+6.9)}} & 75.0 & 74.6\,{\scriptsize\textcolor{red!70!black}{($-$0.4)}} & 75.9\,{\scriptsize\textcolor{green!50!black}{(+0.9)}} & \textbf{76.6}\,{\scriptsize\textcolor{green!50!black}{(+1.6)}}  \\
\midrule
MSVD~\cite{chen2011msvd} & T--V & -- & 70.4 & \textbf{75.8}\,{\scriptsize\textcolor{green!50!black}{(+5.4)}} & 74.3\,{\scriptsize\textcolor{green!50!black}{(+3.9)}} & 73.6\,{\scriptsize\textcolor{green!50!black}{(+3.2)}} & 69.6 & \textbf{75.8}\,{\scriptsize\textcolor{green!50!black}{(+6.2)}} & 75.5\,{\scriptsize\textcolor{green!50!black}{(+5.9)}} & 75.7\,{\scriptsize\textcolor{green!50!black}{(+6.1)}}  \\
\midrule
FLARE~\cite{flare2026} & T--V & -- & 47.4 & 52.1\,{\scriptsize\textcolor{green!50!black}{(+4.7)}} & 52.6\,{\scriptsize\textcolor{green!50!black}{(+5.2)}} & \textbf{54.1}\,{\scriptsize\textcolor{green!50!black}{(+6.7)}} & 52.3 & 53.4\,{\scriptsize\textcolor{green!50!black}{(+1.1)}} & \textbf{53.9}\,{\scriptsize\textcolor{green!50!black}{(+1.6)}} & 53.8\,{\scriptsize\textcolor{green!50!black}{(+1.5)}}  \\
 & T--VA & 56.1 & 46.0 & 52.0\,{\scriptsize\textcolor{green!50!black}{(+6.0)}} & 51.7\,{\scriptsize\textcolor{green!50!black}{(+5.7)}} & \textbf{53.9}\,{\scriptsize\textcolor{green!50!black}{(+7.9)}} & \textbf{52.9} & 52.1\,{\scriptsize\textcolor{red!70!black}{($-$0.8)}} & 52.1\,{\scriptsize\textcolor{red!70!black}{($-$0.8)}} & 52.8\,{\scriptsize\textcolor{red!70!black}{($-$0.1)}}  \\
\midrule
VALOR-32K~\cite{liu2024valor} & T--V & -- & 64.6 & \textbf{71.5}\,{\scriptsize\textcolor{green!50!black}{(+6.9)}} & 70.2\,{\scriptsize\textcolor{green!50!black}{(+5.6)}} & 70.2\,{\scriptsize\textcolor{green!50!black}{(+5.6)}} & 65.6 & \textbf{71.7}\,{\scriptsize\textcolor{green!50!black}{(+6.1)}} & 70.6\,{\scriptsize\textcolor{green!50!black}{(+5.0)}} & 70.4\,{\scriptsize\textcolor{green!50!black}{(+4.8)}}  \\
 & T--VA & 75.9 & 75.2 & 80.5\,{\scriptsize\textcolor{green!50!black}{(+5.3)}} & 80.4\,{\scriptsize\textcolor{green!50!black}{(+5.2)}} & \textbf{80.8}\,{\scriptsize\textcolor{green!50!black}{(+5.6)}} & 75.9 & \textbf{80.9}\,{\scriptsize\textcolor{green!50!black}{(+5.0)}} & 80.5\,{\scriptsize\textcolor{green!50!black}{(+4.6)}} & \textbf{80.9}\,{\scriptsize\textcolor{green!50!black}{(+5.0)}}  \\
\bottomrule
\end{tabular}}
\end{table*}

\begin{table*}[t]
\centering
\caption{Zero-shot video-to-text retrieval (R@1) on eight video benchmarks. For FLARE, T--V uses the vision captions and T--VA the unified captions. The compared methods, evaluation settings and notation are described in the caption of Table~\ref{tab:t2v_wacv}. Bold marks the best result within each backbone group.}
\label{tab:v2t_wacv}\label{tab:v2t_new}
\footnotesize
\setlength{\tabcolsep}{3pt}
\resizebox{0.9\textwidth}{!}{%
\begin{tabular}{@{}llccccccccc@{}}
\toprule
 & & Baseline & \multicolumn{4}{c}{\textsc{Gram} backbone} & \multicolumn{4}{c}{\textsc{HyperGram} backbone} \\
\cmidrule(lr){3-3}\cmidrule(lr){4-7}\cmidrule(lr){8-11}
Dataset & Mode & \textsc{Triangle}$^\star$ & \textsc{Gram}$^\star$ & +\textsc{HyVol} & +\textsc{wHyVol} & +\textsc{wRHG} & \textsc{HyperGram}$^\dag$ & +\textsc{HyVol} & +\textsc{wHyVol} & +\textsc{wRHG}  \\
\midrule
MSR-VTT~\cite{MSRVTT} & T--V & -- & 48.7 & 54.5\,{\scriptsize\textcolor{green!50!black}{(+5.8)}} & 55.1\,{\scriptsize\textcolor{green!50!black}{(+6.4)}} & \textbf{56.5}\,{\scriptsize\textcolor{green!50!black}{(+7.8)}} & 48.7 & 54.2\,{\scriptsize\textcolor{green!50!black}{(+5.5)}} & 55.2\,{\scriptsize\textcolor{green!50!black}{(+6.5)}} & \textbf{55.9}\,{\scriptsize\textcolor{green!50!black}{(+7.2)}}  \\
 & T--VA & 49.7 & 49.3 & 55.8\,{\scriptsize\textcolor{green!50!black}{(+6.5)}} & 56.3\,{\scriptsize\textcolor{green!50!black}{(+7.0)}} & \textbf{57.2}\,{\scriptsize\textcolor{green!50!black}{(+7.9)}} & 49.8 & 56.6\,{\scriptsize\textcolor{green!50!black}{(+6.8)}} & 56.5\,{\scriptsize\textcolor{green!50!black}{(+6.7)}} & \textbf{57.2}\,{\scriptsize\textcolor{green!50!black}{(+7.4)}}  \\
 & T--VAS & -- & 50.1 & \textbf{58.1}\,{\scriptsize\textcolor{green!50!black}{(+8.0)}} & 58.0\,{\scriptsize\textcolor{green!50!black}{(+7.9)}} & 58.0\,{\scriptsize\textcolor{green!50!black}{(+7.9)}} & 50.9 & 59.2\,{\scriptsize\textcolor{green!50!black}{(+8.3)}} & 58.8\,{\scriptsize\textcolor{green!50!black}{(+7.9)}} & \textbf{59.3}\,{\scriptsize\textcolor{green!50!black}{(+8.4)}}  \\
\midrule
DiDeMo~\cite{DIDEMO} & T--V & -- & 49.0 & 53.8\,{\scriptsize\textcolor{green!50!black}{(+4.8)}} & 54.0\,{\scriptsize\textcolor{green!50!black}{(+5.0)}} & \textbf{54.4}\,{\scriptsize\textcolor{green!50!black}{(+5.4)}} & 50.7 & 53.8\,{\scriptsize\textcolor{green!50!black}{(+3.1)}} & 54.6\,{\scriptsize\textcolor{green!50!black}{(+3.9)}} & \textbf{54.8}\,{\scriptsize\textcolor{green!50!black}{(+4.1)}}  \\
 & T--VA & 49.4 & 49.0 & 54.1\,{\scriptsize\textcolor{green!50!black}{(+5.1)}} & \textbf{55.5}\,{\scriptsize\textcolor{green!50!black}{(+6.5)}} & 55.0\,{\scriptsize\textcolor{green!50!black}{(+6.0)}} & 50.8 & 54.7\,{\scriptsize\textcolor{green!50!black}{(+3.9)}} & 55.1\,{\scriptsize\textcolor{green!50!black}{(+4.3)}} & \textbf{55.5}\,{\scriptsize\textcolor{green!50!black}{(+4.7)}}  \\
\midrule
ActivityNet~\cite{ACTIVITYNET} & T--V & -- & 49.3 & 56.2\,{\scriptsize\textcolor{green!50!black}{(+6.9)}} & 61.4\,{\scriptsize\textcolor{green!50!black}{(+12.1)}} & \textbf{61.6}\,{\scriptsize\textcolor{green!50!black}{(+12.3)}} & 50.5 & 55.8\,{\scriptsize\textcolor{green!50!black}{(+5.3)}} & 61.1\,{\scriptsize\textcolor{green!50!black}{(+10.6)}} & \textbf{61.3}\,{\scriptsize\textcolor{green!50!black}{(+10.8)}}  \\
 & T--VA & 52.3 & 49.7 & 56.6\,{\scriptsize\textcolor{green!50!black}{(+6.9)}} & 62.7\,{\scriptsize\textcolor{green!50!black}{(+13.0)}} & \textbf{63.0}\,{\scriptsize\textcolor{green!50!black}{(+13.3)}} & 50.9 & 56.4\,{\scriptsize\textcolor{green!50!black}{(+5.5)}} & 62.8\,{\scriptsize\textcolor{green!50!black}{(+11.9)}} & \textbf{63.0}\,{\scriptsize\textcolor{green!50!black}{(+12.1)}}  \\
\midrule
VATEX~\cite{wang2019vatex} & T--V & -- & 84.7 & 92.3\,{\scriptsize\textcolor{green!50!black}{(+7.6)}} & \textbf{92.6}\,{\scriptsize\textcolor{green!50!black}{(+7.9)}} & \textbf{92.6}\,{\scriptsize\textcolor{green!50!black}{(+7.9)}} & 86.3 & \textbf{92.4}\,{\scriptsize\textcolor{green!50!black}{(+6.1)}} & 92.3\,{\scriptsize\textcolor{green!50!black}{(+6.0)}} & 92.1\,{\scriptsize\textcolor{green!50!black}{(+5.8)}}  \\
 & T--VA & 86.5 & 86.5 & 92.8\,{\scriptsize\textcolor{green!50!black}{(+6.3)}} & \textbf{94.0}\,{\scriptsize\textcolor{green!50!black}{(+7.5)}} & 93.7\,{\scriptsize\textcolor{green!50!black}{(+7.2)}} & 88.4 & 92.9\,{\scriptsize\textcolor{green!50!black}{(+4.5)}} & \textbf{93.5}\,{\scriptsize\textcolor{green!50!black}{(+5.1)}} & 93.3\,{\scriptsize\textcolor{green!50!black}{(+4.9)}}  \\
 & T--VAS & -- & 86.5 & 93.8\,{\scriptsize\textcolor{green!50!black}{(+7.3)}} & \textbf{94.4}\,{\scriptsize\textcolor{green!50!black}{(+7.9)}} & \textbf{94.4}\,{\scriptsize\textcolor{green!50!black}{(+7.9)}} & 87.8 & 93.9\,{\scriptsize\textcolor{green!50!black}{(+6.1)}} & \textbf{94.0}\,{\scriptsize\textcolor{green!50!black}{(+6.2)}} & \textbf{94.0}\,{\scriptsize\textcolor{green!50!black}{(+6.2)}}  \\
\midrule
Shot2Story~\cite{han2023shot2story} & T--V & -- & 66.4 & \textbf{77.6}\,{\scriptsize\textcolor{green!50!black}{(+11.2)}} & 75.7\,{\scriptsize\textcolor{green!50!black}{(+9.3)}} & 76.4\,{\scriptsize\textcolor{green!50!black}{(+10.0)}} & 73.0 & \textbf{78.1}\,{\scriptsize\textcolor{green!50!black}{(+5.1)}} & 75.9\,{\scriptsize\textcolor{green!50!black}{(+2.9)}} & 75.8\,{\scriptsize\textcolor{green!50!black}{(+2.8)}}  \\
 & T--VA & 74.8 & 66.2 & \textbf{77.6}\,{\scriptsize\textcolor{green!50!black}{(+11.4)}} & 75.7\,{\scriptsize\textcolor{green!50!black}{(+9.5)}} & 76.3\,{\scriptsize\textcolor{green!50!black}{(+10.1)}} & 73.1 & \textbf{78.1}\,{\scriptsize\textcolor{green!50!black}{(+5.0)}} & 76.2\,{\scriptsize\textcolor{green!50!black}{(+3.1)}} & 76.0\,{\scriptsize\textcolor{green!50!black}{(+2.9)}}  \\
\midrule
MSVD~\cite{chen2011msvd} & T--V & -- & 69.0 & 74.5\,{\scriptsize\textcolor{green!50!black}{(+5.5)}} & \textbf{74.6}\,{\scriptsize\textcolor{green!50!black}{(+5.6)}} & 74.5\,{\scriptsize\textcolor{green!50!black}{(+5.5)}} & 67.0 & 74.6\,{\scriptsize\textcolor{green!50!black}{(+7.6)}} & 74.6\,{\scriptsize\textcolor{green!50!black}{(+7.6)}} & \textbf{75.2}\,{\scriptsize\textcolor{green!50!black}{(+8.2)}}  \\
\midrule
FLARE~\cite{flare2026} & T--V & -- & 47.8 & 55.6\,{\scriptsize\textcolor{green!50!black}{(+7.8)}} & 55.7\,{\scriptsize\textcolor{green!50!black}{(+7.9)}} & \textbf{55.8}\,{\scriptsize\textcolor{green!50!black}{(+8.0)}} & 52.4 & \textbf{56.4}\,{\scriptsize\textcolor{green!50!black}{(+4.0)}} & \textbf{56.4}\,{\scriptsize\textcolor{green!50!black}{(+4.0)}} & 55.9\,{\scriptsize\textcolor{green!50!black}{(+3.5)}}  \\
 & T--VA & 55.0 & 46.8 & \textbf{55.5}\,{\scriptsize\textcolor{green!50!black}{(+8.7)}} & 54.6\,{\scriptsize\textcolor{green!50!black}{(+7.8)}} & 55.0\,{\scriptsize\textcolor{green!50!black}{(+8.2)}} & 52.6 & \textbf{56.4}\,{\scriptsize\textcolor{green!50!black}{(+3.8)}} & 55.9\,{\scriptsize\textcolor{green!50!black}{(+3.3)}} & 55.8\,{\scriptsize\textcolor{green!50!black}{(+3.2)}}  \\
\midrule
VALOR-32K~\cite{liu2024valor} & T--V & -- & 61.4 & \textbf{71.8}\,{\scriptsize\textcolor{green!50!black}{(+10.4)}} & 69.9\,{\scriptsize\textcolor{green!50!black}{(+8.5)}} & 70.2\,{\scriptsize\textcolor{green!50!black}{(+8.8)}} & 62.5 & \textbf{72.5}\,{\scriptsize\textcolor{green!50!black}{(+10.0)}} & 71.4\,{\scriptsize\textcolor{green!50!black}{(+8.9)}} & 71.5\,{\scriptsize\textcolor{green!50!black}{(+9.0)}}  \\
 & T--VA & 74.7 & 71.2 & \textbf{81.6}\,{\scriptsize\textcolor{green!50!black}{(+10.4)}} & 81.1\,{\scriptsize\textcolor{green!50!black}{(+9.9)}} & 80.8\,{\scriptsize\textcolor{green!50!black}{(+9.6)}} & 72.4 & \textbf{82.2}\,{\scriptsize\textcolor{green!50!black}{(+9.8)}} & 82.0\,{\scriptsize\textcolor{green!50!black}{(+9.6)}} & 81.8\,{\scriptsize\textcolor{green!50!black}{(+9.4)}}  \\
\bottomrule
\end{tabular}}
\end{table*}

\subsubsection{Audio--text retrieval}
\label{sec:audio_results}

Table~\ref{tab:audio_ret} shows that refinement also benefits audio--text retrieval, though the gains are less uniform than on the video benchmarks. On AudioCaps, all three configurations improve over their corresponding backbones in both retrieval directions. \hypergram+\why yields the strongest results, increasing text-to-audio R@1 from 32.4 to 36.9 and audio-to-text R@1 from 33.2 to 37.5. \hypergram+\wrhg reaches 36.8 and 36.2, respectively, indicating that relational propagation offers no additional advantage over weighted additive propagation in this setting.

Clotho reveals a stronger dependence on the configuration and retrieval direction. For text-to-audio retrieval, \hy~\cite{nag2027hyvol} reduces R@1 from 11.9 to 11.0 with \gram and from 12.9 to 11.5 with \hypergram. The weighted configurations recover part or all of this loss: \gram+\why reaches 13.2, while \hypergram+\wrhg reaches 13.0. In audio-to-text retrieval, \why achieves the highest score for both backbones, increasing R@1 from 11.5 to 15.0 with \gram and from 12.1 to 13.7 with \hypergram.

On SoundDescs, improvements are small by comparison, and \hy achieves the highest scores among the refinement configurations for both backbones and retrieval directions. For example, \gram+\hy increases text-to-audio R@1 from 3.6 to 4.6, while \hypergram+\hy increases it from 3.9 to 4.5. The low absolute recall indicates that this benchmark remains challenging under the evaluated training and retrieval protocol.

Unlike AudioCaps, Clotho and SoundDescs contain only audio on the candidate side. Their results therefore assess transfer to unimodal candidates rather than the direct benefit of combining multiple streams at inference. They demonstrate that training-time refinement can benefit audio-only retrieval, while also showing that the more elaborate configurations are not consistently preferable in this setting.

\subsubsection{Comparison with existing methods}
\label{sec:external_comparisons}

Table~\ref{tab:published} compares video retrieval results with those of existing methods, grouped by available candidate modalities. Both \gram+\wrhg and \hypergram+\wrhg outperform the re-evaluated \pmrl checkpoint~\cite{liu2026principled} in every reported setting in this table. On ActivityNet T--VA, for example, \pmrl obtains 54.1 for text-to-video retrieval and 48.6 for video-to-text retrieval, compared with 60.9 and 63.0 for \hypergram+\wrhg. On MSR-VTT T--VAS, \hypergram+\wrhg achieves 57.3 and 59.3, compared with 54.3 and 52.9 for \pmrl.

The comparison with \tri~\cite{cicchetti2026triangle} is more mixed. On VALOR-32K T--VA, \gram+\wrhg reaches 80.8 for text-to-video retrieval, compared with 75.9 for \tri. Conversely, \tri obtains 56.1 on FLARE T--VA, exceeding the corresponding scores of 53.9 and 52.8 for \gram+\wrhg and \hypergram+\wrhg. On AudioCaps, \tri also remains stronger than the refinement configurations, achieving 37.4 for text-to-audio retrieval and 38.2 for audio-to-text retrieval.

Table~\ref{tab:published_audio} provides additional context for AudioCaps text-to-audio retrieval. In the T--AV setting, \gram+\wrhg and \hypergram+\wrhg achieve 35.4 and 36.8, respectively, compared with 34.4 for the re-evaluated \pmrl checkpoint. Published results obtained with other pretraining corpora or evaluation procedures are included as reference values. They establish the broader performance context but do not constitute controlled comparisons of fusion mechanisms.

\subsubsection{Neighbourhood-Size Sensitivity}
\label{sec:neighborhood}
The neighbourhood size $k$ in Eq.~\eqref{eq:connectivity} determines how many caption neighbours each candidate considers before the mutual-neighbour test is applied. Table~\ref{tab:topk_ablation} varies $k$ over $\{4,8,12,16\}$ for \gram+\hy while keeping all other settings fixed. Since each GPU builds its hypergraph from 64 clips, $k_{\mathrm{eff}}=\min(k,16)$, and $k=16$ is the largest neighbourhood a batch can support.

Increasing $k$ from 4 to 12 improves R@1 in every setting, and the effect is most pronounced for video-to-text retrieval. Averaged over the modality sets of each benchmark, the video-to-text gain is 6.6 points on MSR-VTT, 5.8 on DiDeMo, and 8.3 on ActivityNet; on VATEX, where R@1 already exceeds 90, it is 0.8 points. The corresponding text-to-video gains are smaller but consistent, for example from 53.2 to 55.8 on MSR-VTT T--VAS and from 50.3 to 54.6 on DiDeMo T--VA. A further increase to $k=16$ brings no average benefit and markedly degrades video-to-text retrieval on ActivityNet, where R@1 falls from 56.2 to 49.2 in T--V and from 56.6 to 49.8 in T--VA. Across the 20 settings, $k=12$ yields the best or tied-best result in 19; the only exception is VATEX T--VA text-to-video retrieval, where $k=16$ is 0.3 points higher. We therefore adopt $k=12$ for all configurations and benchmarks.

\begin{table}[t]
\centering
\caption{Sensitivity to the number of semantic neighbours $k$. Values are R@1 for text-to-video (T$\rightarrow$V) and video-to-text (V$\rightarrow$T) retrieval; bold marks the best tested $k$ in each row, and $k{=}12$ is used in all main experiments. The reported configuration uses \textsc{\hy} on top of the \textsc{\gram} backbone.}
\label{tab:topk_ablation}
\footnotesize
\setlength{\tabcolsep}{2.5pt}
\resizebox{\columnwidth}{!}{%
\begin{tabular}{@{}llcccccccc@{}}
\toprule
& & \multicolumn{4}{c}{T$\rightarrow$V} & \multicolumn{4}{c}{V$\rightarrow$T} \\
\cmidrule(lr){3-6} \cmidrule(l){7-10}
Dataset & Mode & $k{=}4$ & $k{=}8$ & $k{=}12$ & $k{=}16$ & $k{=}4$ & $k{=}8$ & $k{=}12$ & $k{=}16$ \\
\midrule
MSR-VTT~\cite{MSRVTT} & T--V   & 50.0 & 52.1 & \textbf{52.6} & 51.1 & 47.7 & 52.2 & \textbf{54.5} & 54.4 \\
                      & T--VA  & 50.6 & 52.5 & \textbf{54.0} & 52.1 & 49.0 & 51.8 & \textbf{55.8} & 55.7 \\
                      & T--VAS & 53.2 & 54.3 & \textbf{55.8} & 54.7 & 51.9 & 53.7 & \textbf{58.1} & 57.6 \\
\midrule
DiDeMo~\cite{DIDEMO}  & T--V   & 50.0 & 51.8 & \textbf{53.8} & 52.5 & 48.0 & 52.6 & \textbf{53.8} & \textbf{53.8} \\
                      & T--VA  & 50.3 & 51.4 & \textbf{54.6} & 52.4 & 48.4 & 52.7 & \textbf{54.1} & 53.3 \\
\midrule
ActivityNet~\cite{ACTIVITYNET} & T--V  & 53.3 & 55.9 & \textbf{56.2} & 54.9 & 47.8 & 50.6 & \textbf{56.2} & 49.2 \\
                               & T--VA & 53.9 & 56.7 & \textbf{56.9} & 55.4 & 48.4 & 50.2 & \textbf{56.6} & 49.8 \\
\midrule
VATEX~\cite{wang2019vatex} & T--V   & 87.7 & 91.9 & \textbf{92.1} & 91.9 & 91.4 & 92.1 & \textbf{92.3} & 91.6 \\
                           & T--VA  & 89.0 & 91.9 & 93.0 & \textbf{93.3} & 91.7 & 92.1 & \textbf{92.8} & 92.3 \\
                           & T--VAS & 90.1 & 92.6 & \textbf{94.0} & 93.5 & 93.3 & 93.7 & \textbf{93.8} & 93.0 \\
\bottomrule
\end{tabular}}
\end{table}

\begin{table*}[t]
\centering
\caption{Zero-shot text-to-audio and audio-to-text retrieval (R@1). The compared methods, evaluation settings and notation are described in the caption of Table~\ref{tab:t2v_wacv}. Bold marks the best result within each backbone group.}
\label{tab:audio_ret}
\footnotesize
\setlength{\tabcolsep}{3pt}
\resizebox{0.9\textwidth}{!}{%
\begin{tabular}{@{}llccccccccc@{}}
\toprule
 & & Baseline & \multicolumn{4}{c}{\textsc{Gram} backbone} & \multicolumn{4}{c}{\textsc{HyperGram} backbone} \\
\cmidrule(lr){3-3}\cmidrule(lr){4-7}\cmidrule(lr){8-11}
Dataset & Mode & \textsc{Triangle}$^\star$ & \textsc{Gram}$^\star$ & +\textsc{HyVol} & +\textsc{wHyVol} & +\textsc{wRHG} & \textsc{HyperGram}$^\dag$ & +\textsc{HyVol} & +\textsc{wHyVol} & +\textsc{wRHG} \\
\midrule
\multicolumn{11}{c}{\textit{Text-to-audio}} \\
\midrule
AudioCaps~\cite{AUDIOCAPS} & T--AV & 37.4 & 33.4 & \textbf{35.7}\,{\scriptsize\textcolor{green!50!black}{(+2.3)}} & 35.2\,{\scriptsize\textcolor{green!50!black}{(+1.8)}} & 35.4\,{\scriptsize\textcolor{green!50!black}{(+2.0)}} & 32.4 & 36.1\,{\scriptsize\textcolor{green!50!black}{(+3.7)}} & \textbf{36.9}\,{\scriptsize\textcolor{green!50!black}{(+4.5)}} & 36.8\,{\scriptsize\textcolor{green!50!black}{(+4.4)}} \\
Clotho~\cite{drossos2020clotho} & T--A & -- & 11.9 & 11.0\,{\scriptsize\textcolor{red!70!black}{($-$0.9)}} & \textbf{13.2}\,{\scriptsize\textcolor{green!50!black}{(+1.3)}} & 12.3\,{\scriptsize\textcolor{green!50!black}{(+0.4)}} & 12.9 & 11.5\,{\scriptsize\textcolor{red!70!black}{($-$1.4)}} & 12.3\,{\scriptsize\textcolor{red!70!black}{($-$0.6)}} & \textbf{13.0}\,{\scriptsize\textcolor{green!50!black}{(+0.1)}} \\
SoundDescs~\cite{koepke2023audio} & T--A & -- & 3.6 & \textbf{4.6}\,{\scriptsize\textcolor{green!50!black}{(+1.0)}} & 4.5\,{\scriptsize\textcolor{green!50!black}{(+0.9)}} & 4.5\,{\scriptsize\textcolor{green!50!black}{(+0.9)}} & 3.9 & \textbf{4.5}\,{\scriptsize\textcolor{green!50!black}{(+0.6)}} & 4.2\,{\scriptsize\textcolor{green!50!black}{(+0.3)}} & 4.1\,{\scriptsize\textcolor{green!50!black}{(+0.2)}} \\
\midrule
\multicolumn{11}{c}{\textit{Audio-to-text}} \\
\midrule
AudioCaps~\cite{AUDIOCAPS} & T--AV & 38.2 & 33.4 & 33.7\,{\scriptsize\textcolor{green!50!black}{(+0.3)}} & 33.8\,{\scriptsize\textcolor{green!50!black}{(+0.4)}} & \textbf{34.9}\,{\scriptsize\textcolor{green!50!black}{(+1.5)}} & 33.2 & 36.4\,{\scriptsize\textcolor{green!50!black}{(+3.2)}} & \textbf{37.5}\,{\scriptsize\textcolor{green!50!black}{(+4.3)}} & 36.2\,{\scriptsize\textcolor{green!50!black}{(+3.0)}} \\
Clotho~\cite{drossos2020clotho} & T--A & -- & 11.5 & 12.5\,{\scriptsize\textcolor{green!50!black}{(+1.0)}} & \textbf{15.0}\,{\scriptsize\textcolor{green!50!black}{(+3.5)}} & 14.0\,{\scriptsize\textcolor{green!50!black}{(+2.5)}} & 12.1 & 12.4\,{\scriptsize\textcolor{green!50!black}{(+0.3)}} & \textbf{13.7}\,{\scriptsize\textcolor{green!50!black}{(+1.6)}} & 13.6\,{\scriptsize\textcolor{green!50!black}{(+1.5)}} \\
SoundDescs~\cite{koepke2023audio} & T--A & -- & 4.3 & \textbf{4.7}\,{\scriptsize\textcolor{green!50!black}{(+0.4)}} & 4.5\,{\scriptsize\textcolor{green!50!black}{(+0.2)}} & 4.3\,{\scriptsize\textcolor{gray}{(0.0)}} & 4.2 & \textbf{4.6}\,{\scriptsize\textcolor{green!50!black}{(+0.4)}} & 4.4\,{\scriptsize\textcolor{green!50!black}{(+0.2)}} & 4.3\,{\scriptsize\textcolor{green!50!black}{(+0.1)}} \\
\bottomrule
\end{tabular}}
\end{table*}

\begin{table*}[t]
\centering
\caption{Comparison with published zero-shot retrieval results (R@1) on the four benchmarks for which such results exist. Rows are grouped by modality set, and methods are compared only within the same group. $^\S$Numbers reported in the original papers under their own pretraining data and protocols (reference only). $^\star$Authors' released checkpoint, evaluated by us on the same test data as our models. A dash (--) marks a result that is not available. Bold marks the best result within each modality group.}
\label{tab:published}
\footnotesize
\resizebox{1.7\columnwidth}{!}{%
\begin{tabular}{@{}>{\raggedright\arraybackslash}p{3.8cm}*{8}{>{\centering\arraybackslash}p{\dimexpr(\textwidth-3.8cm-16\tabcolsep)/8\relax}}@{}}
\toprule
 & \multicolumn{2}{c}{MSR-VTT~\cite{MSRVTT}} & \multicolumn{2}{c}{DiDeMo~\cite{DIDEMO}} & \multicolumn{2}{c}{ActivityNet~\cite{ACTIVITYNET}} & \multicolumn{2}{c}{VATEX~\cite{wang2019vatex}} \\
\cmidrule(lr){2-3}\cmidrule(lr){4-5}\cmidrule(lr){6-7}\cmidrule(lr){8-9}
Method & T$\to$V & V$\to$T & T$\to$V & V$\to$T & T$\to$V & V$\to$T & T$\to$V & V$\to$T \\
\midrule
\multicolumn{9}{@{}l}{\emph{Text--video (T--V)}} \\
\midrule
HiTeA$^\S$~\cite{Ye2022HiTeAHT} & 34.4 & 46.8 & 43.2 & \textbf{56.5} & -- & -- & -- & -- \\
ImageBind$^\S$~\cite{Girdhar2023ImageBindOE} & 36.8 & 36.8 & -- & -- & -- & -- & -- & -- \\
InternVideo-L$^\S$~\cite{Wang2022InternVideoGV} & 40.7 & 39.6 & 31.5 & 33.5 & 30.7 & 31.4 & 49.5 & 69.5 \\
LanguageBind$^\S$~\cite{zhu2024languagebind} & 44.8 & 40.9 & 39.9 & 39.8 & 41.0 & 39.1 & -- & -- \\
mPLUG-2$^\S$~\cite{Xu2023mPLUG2AM} & 47.1 & -- & 45.7 & -- & -- & -- & -- & -- \\
Norton$^\S$~\cite{lin2024norton} & 10.7 & -- & -- & -- & -- & -- & -- & -- \\
OmniVL$^\S$~\cite{Wang2022OmniVLOF} & 42.0 & 34.6 & 40.6 & 33.3 & -- & -- & -- & -- \\
TVTSv2$^\S$~\cite{Zeng2023TVTSv2LO} & 38.2 & 38.2 & 34.6 & 34.6 & -- & -- & -- & -- \\
UMT$^\S$~\cite{Li2023UnmaskedTT} & 33.3 & 33.3 & 34.0 & 34.0 & 31.9 & 31.9 & -- & -- \\
UMT-L$^\S$~\cite{Li2023UnmaskedTT} & 40.7 & 40.7 & 48.6 & 24.9 & 41.9 & 39.4 & -- & -- \\
ViCLIP$^\S$~\cite{Wang2023InternVidAL} & 42.4 & 41.3 & 18.4 & 27.9 & 15.1 & 24.0 & -- & -- \\
VideoCoCa$^\S$~\cite{yan2022videococa} & 34.3 & 34.3 & -- & -- & 34.5 & 34.5 & 53.2 & -- \\
VideoPrism-b$^\S$~\cite{Zhao2024VideoPrismAF} & 51.4 & 50.2 & -- & -- & 49.6 & 47.9 & 62.5 & 76.2 \\
\addlinespace[2pt]
\textsc{PMRL}$^\star$~\cite{liu2026principled} & 51.3 & 49.6 & 53.0 & 50.5 & 53.8 & 48.0 & 87.7 & 85.4 \\
\rowcolor{gray!15}
\textsc{Gram}+\textsc{wRHG} (ours) & 53.0 & \textbf{56.5} & \textbf{54.0} & 54.4 & 59.6 & \textbf{61.6} & \textbf{92.3} & \textbf{92.6} \\
\rowcolor{gray!15}
\textsc{HyperGram}+\textsc{wRHG} (ours) & \textbf{54.2} & 55.9 & \textbf{54.0} & 54.8 & \textbf{59.9} & 61.3 & \textbf{92.3} & 92.1 \\
\midrule
\multicolumn{9}{@{}l}{\emph{Text--video--audio (T--VA)}} \\
\midrule
VAST$^\S$~\cite{chen2023vast} & 49.3 & 43.7 & 49.5 & 48.2 & 51.4 & 46.8 & 80.0 & 77.1 \\
\addlinespace[2pt]
\textsc{PMRL}$^\star$~\cite{liu2026principled} & 50.9 & 51.0 & 52.5 & 49.3 & 54.1 & 48.6 & 88.9 & 86.1 \\
\rowcolor{gray!15}
\textsc{Gram}+\textsc{wRHG} (ours) & 54.7 & \textbf{57.2} & 54.1 & 55.0 & 60.7 & \textbf{63.0} & \textbf{93.5} & \textbf{93.7} \\
\rowcolor{gray!15}
\textsc{HyperGram}+\textsc{wRHG} (ours) & \textbf{55.1} & \textbf{57.2} & \textbf{54.6} & \textbf{55.5} & \textbf{60.9} & \textbf{63.0} & \textbf{93.5} & 93.3 \\
\midrule
\multicolumn{9}{@{}l}{\emph{Text--video--audio--subtitle (T--VAS)}} \\
\midrule
VAST$^\S$~\cite{chen2023vast} & 50.7 & 49.0 & -- & -- & -- & -- & 82.1 & 78.7 \\
\addlinespace[2pt]
\textsc{PMRL}$^\star$~\cite{liu2026principled} & 54.3 & 52.9 & -- & -- & -- & -- & 89.6 & 87.0 \\
\rowcolor{gray!15}
\textsc{Gram}+\textsc{wRHG} (ours) & \textbf{57.4} & 58.0 & -- & -- & -- & -- & \textbf{94.0} & \textbf{94.4} \\
\rowcolor{gray!15}
\textsc{HyperGram}+\textsc{wRHG} (ours) & 57.3 & \textbf{59.3} & -- & -- & -- & -- & 93.5 & 94.0 \\
\bottomrule
\end{tabular}}
\end{table*}

\begin{table}[t]
\centering
\caption{Comparison with published text-to-audio retrieval results (R@1) on AudioCaps~\cite{AUDIOCAPS}. $^\S$Published numbers (reference only); $^\star$released checkpoint evaluated by us.}
\label{tab:published_audio}
\footnotesize
\setlength{\tabcolsep}{8pt}
\begin{tabular}{@{}lcc@{}}
\toprule
Method & T--A & T--AV \\
\midrule
AVFIC$^\S$~\cite{Nagrani2022LearningAM} & 8.7 & 10.6 \\
ImageBind$^\S$~\cite{Girdhar2023ImageBindOE} & 9.3 & -- \\
LanguageBind$^\S$~\cite{zhu2024languagebind} & 19.7 & -- \\
VIP-ANT$^\S$~\cite{Zhao2021ConnectingTD} & 27.7 & -- \\
VAST$^\S$~\cite{chen2023vast} & -- & 32.1 \\
\addlinespace[2pt]
\textsc{PMRL}$^\star$~\cite{liu2026principled} & -- & 34.4 \\
\rowcolor{gray!15}
\textsc{Gram}+\textsc{wRHG} (ours) & -- & 35.4 \\
\rowcolor{gray!15}
\textsc{HyperGram}+\textsc{wRHG} (ours) & -- & \textbf{36.8} \\
\bottomrule
\end{tabular}
\end{table}

\begin{figure*}[t]
\centering
\includegraphics[width=0.7\textwidth]{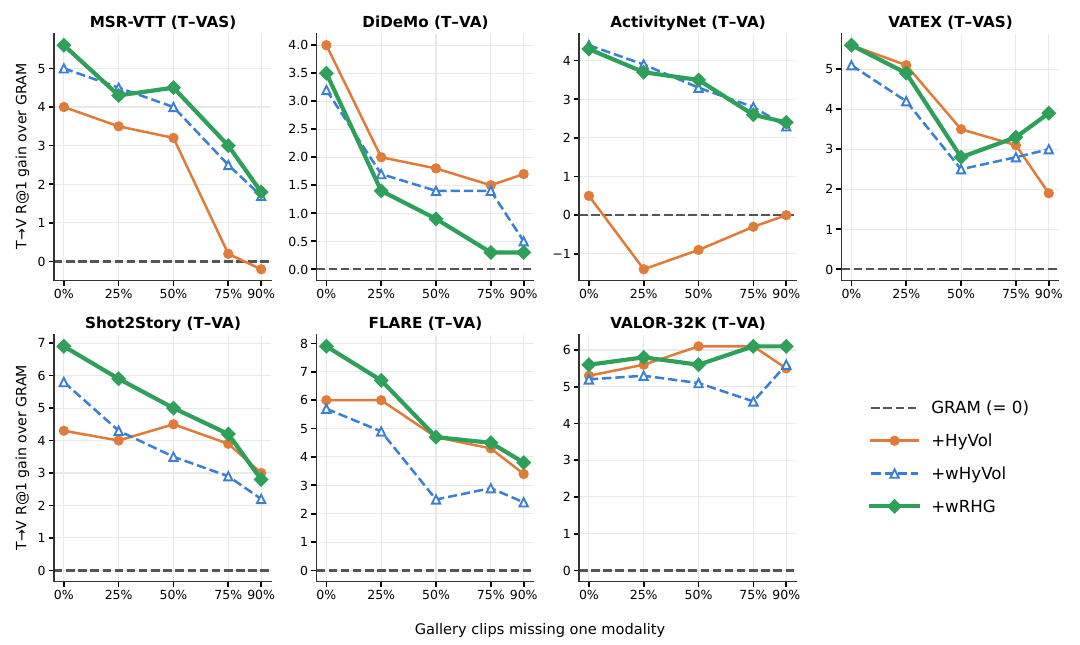}
\caption{Zero-shot text-to-video retrieval under missing modalities with the \textsc{\gram} backbone on seven video benchmarks. Each panel shows the R@1 gain of +\textsc{\hy}, +\textsc{\why} and +\textsc{\wrhg} over \textsc{\gram} (dashed line at zero) as a fraction $r$ of the gallery clips that lose one randomly chosen modality; the per-clip masks are fixed and identical for all methods, and $r{=}0$ is the standard setting. MSR-VTT and VATEX use the T--VAS modality set, and the other benchmarks use T--VA.}
\label{fig:miss_t2v_gram}
\end{figure*}

\begin{figure*}[t]
\centering
\includegraphics[width=0.8\textwidth]{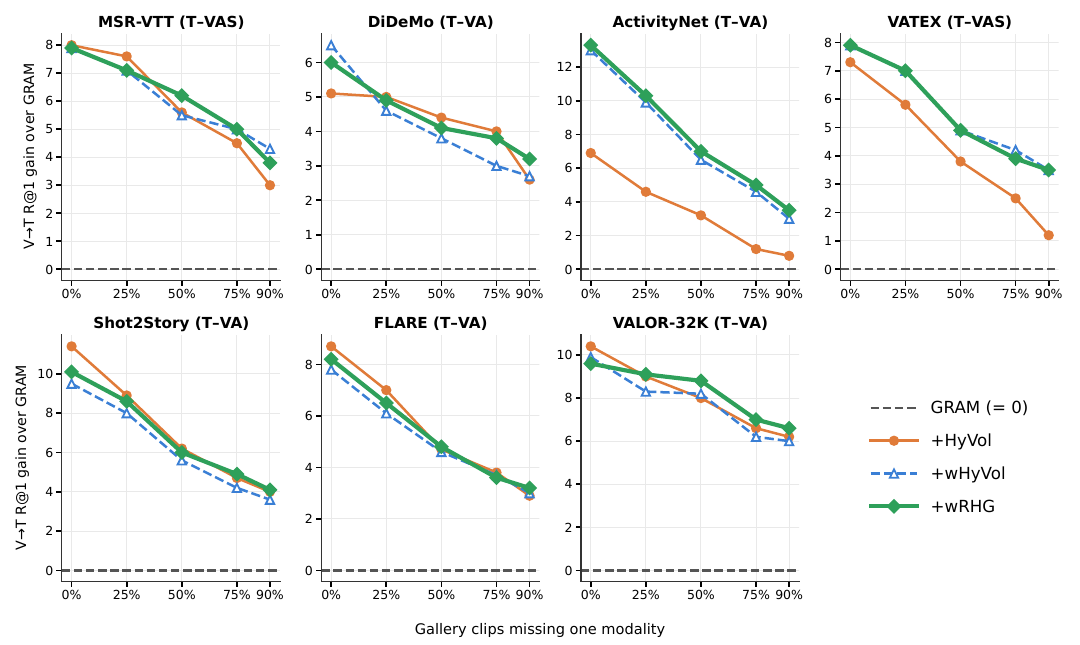}
\caption{Zero-shot video-to-text retrieval under missing modalities with the \textsc{\gram} backbone: R@1 gain over \textsc{\gram} (dashed line at zero). The protocol is the same as in Figure~\ref{fig:miss_t2v_gram}.}
\label{fig:miss_v2t_gram}
\end{figure*}

\begin{figure*}[t]
\centering
\includegraphics[width=0.8\textwidth]{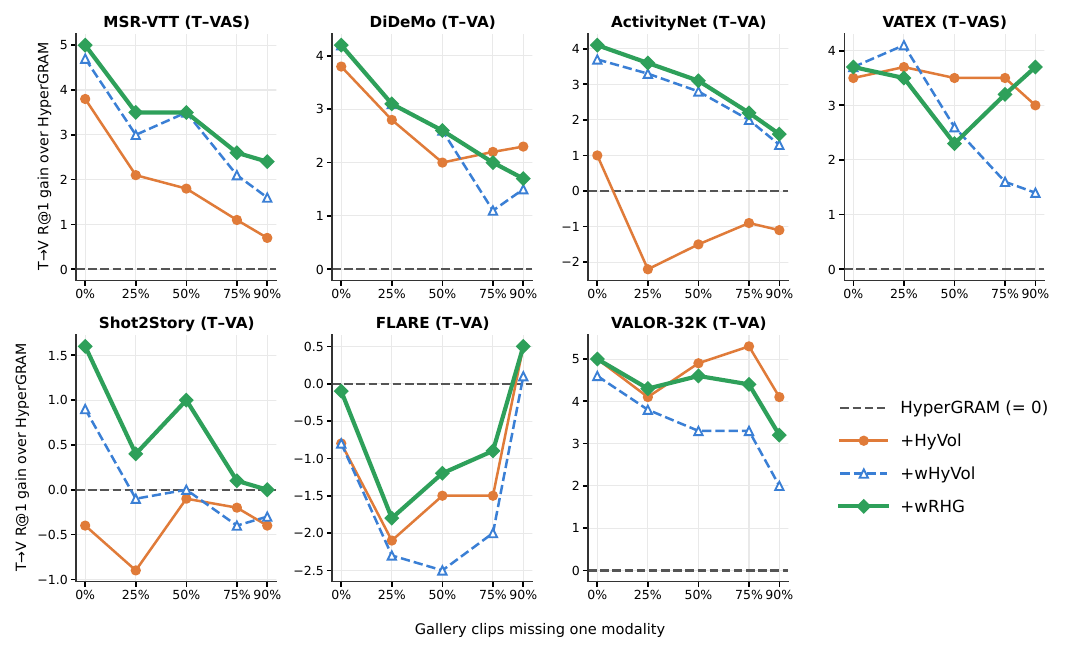}
\caption{Zero-shot text-to-video retrieval under missing modalities with the \textsc{\hypergram} backbone: R@1 gain over \textsc{\hypergram} (dashed line at zero). The protocol is the same as in Figure~\ref{fig:miss_t2v_gram}.}
\label{fig:miss_t2v_hypergram}
\end{figure*}

\begin{figure*}[t]
\centering
\includegraphics[width=0.8\textwidth]{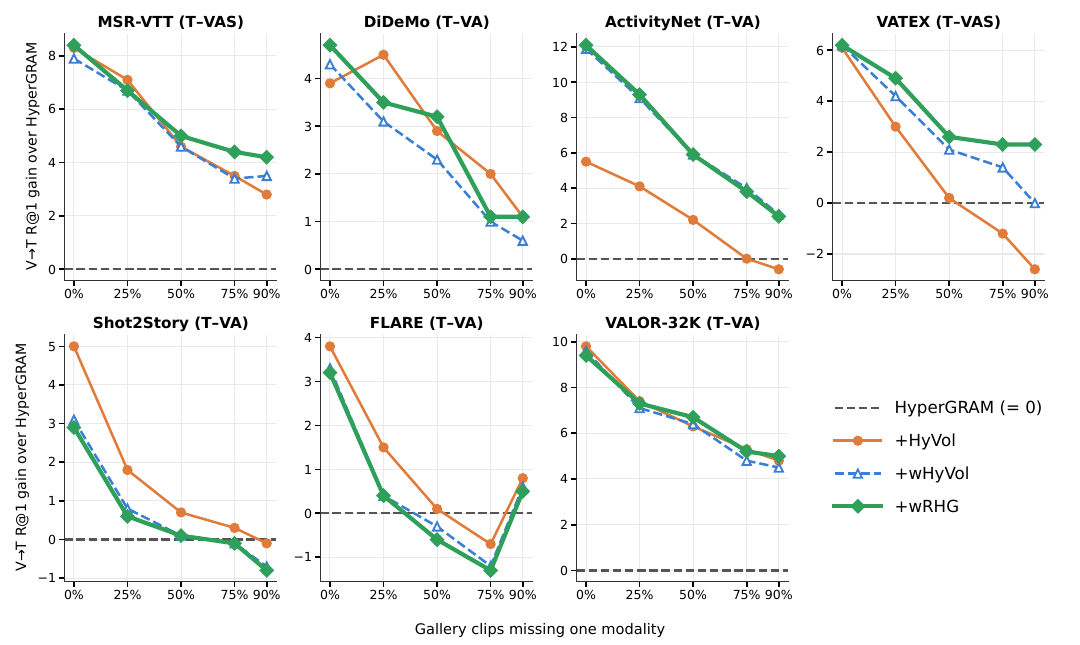}
\caption{Zero-shot video-to-text retrieval under missing modalities with the \textsc{\hypergram} backbone: R@1 gain over \textsc{\hypergram} (dashed line at zero). The protocol is the same as in Figure~\ref{fig:miss_t2v_gram}.}
\label{fig:miss_v2t_hypergram}
\end{figure*}

\subsection{Robustness to Missing Modalities}
\label{sec:missing_modalities}

We further evaluate retrieval when individual clips lose part of their multimodal information. Unlike the complete-modality comparisons, this setting introduces heterogeneous observations within the same collection: some clips retain all available streams, while others are represented by a reduced modality set. Figures~\ref{fig:miss_t2v_gram}--\ref{fig:miss_v2t_hypergram} examine whether improvements obtained through training-time hypergraph refinement persist under these conditions.

\subsubsection{Evaluation protocol}
\label{sec:missing_protocol}

The evaluation covers seven video benchmarks: MSR-VTT, DiDeMo, ActivityNet, VATEX, Shot2Story, FLARE, and VALOR-32K. MSR-VTT and VATEX use the T--VAS setting, while the remaining benchmarks use T--VA. MSVD is excluded because its evaluation configuration contains only video.

For a fraction $r$ of the clips, one randomly selected modality is removed. We consider missingness rates of 0\%, 25\%, 50\%, 75\%, and 90\%, with 0\% corresponding to the standard evaluation setting. The per-clip masks are fixed and identical across methods, ensuring that performance differences are measured under the same availability conditions. Each affected clip retains two modalities in T--VAS and one modality in T--VA. Thus, the missingness rate specifies the proportion of incomplete clips rather than the proportion of all modality observations removed.

The affected clips serve as gallery items in text-to-video retrieval and as queries in video-to-text retrieval. Figures~\ref{fig:miss_t2v_gram} and~\ref{fig:miss_v2t_gram} report results with \gram, while Figures~\ref{fig:miss_t2v_hypergram} and~\ref{fig:miss_v2t_hypergram} report results with \hypergram. Each curve shows the difference in R@1 between a refinement configuration and its corresponding backbone at the same missingness rate. Positive values therefore indicate an advantage over the backbone under incomplete observations; they do not imply that absolute retrieval performance is unchanged as modalities become unavailable. Table~\ref{tab:miss_all} in Appendix~\ref{app:additional} lists the underlying R@1 values for each method compared at each missingness rate.

\subsubsection{Results with \gram}
\label{sec:missing_gram}

Figure~\ref{fig:miss_t2v_gram} shows that \why and \wrhg retain positive text-to-video gains over \gram on all seven benchmarks at every evaluated missingness rate. Their advantage therefore persists even when 90\% of the clips lose one modality. Although the gains generally become smaller as missingness increases, the weighted configurations remain beneficial across the evaluated collections.

ActivityNet clearly distinguishes the weighted configurations from \hy~\cite{nag2027hyvol}. Without missing modalities, the T--VA gains over \gram are 4.4 percentage points for \why and 4.3 points for \wrhg, compared with $0.5$ points for \hy. As shown in Figure~\ref{fig:miss_t2v_gram}, both weighted configurations remain above the backbone throughout the missingness sweep, whereas \hy falls below it at $25$\%, $50$\%, and $75$\% missingness. The weighted configurations therefore maintain their advantage as audiovisual observations become incomplete.

A similar distinction appears on MSR-VTT. In the complete T--VAS setting, \why and \wrhg improve over \gram by 5.0 and 5.6 points, respectively, compared with 4.0 points for \hy. At 90\% missingness, both weighted configurations remain positive, whereas \hy falls slightly below the backbone. On VATEX, the weighted configurations also retain a larger advantage than \hy at the highest missingness rate.

This ordering does not hold uniformly. On DiDeMo, \hy provides a larger text-to-video gain than either weighted configuration at the nonzero missingness rates. On VALOR-$32$K, all three configurations remain above \gram throughout the sweep, and their gains do not decrease monotonically. These differences indicate that the relative benefit of weighting depends on the benchmark as well as the amount of missing information.

For video-to-text retrieval, Figure~\ref{fig:miss_v2t_gram} shows positive gains for all three refinement configurations across all evaluated missingness rates. The gains generally narrow as more clips become incomplete. On ActivityNet, the complete-modality improvements of 13.0 points for \why and 13.3 points for \wrhg decrease with missingness, but both configurations remain above \hy throughout the sweep. On VATEX, the weighted configurations likewise retain a clearer advantage at high missingness rates. Overall, the \gram results show that refinement remains beneficial under incomplete observations, while the magnitude and ordering of the improvements vary across datasets.

\subsubsection{Results with \hypergram}
\label{sec:missing_hypergram}

The \hypergram results in Figures~\ref{fig:miss_t2v_hypergram} and~\ref{fig:miss_v2t_hypergram} vary more across benchmarks. On MSR-VTT and ActivityNet, both weighted configurations retain positive gains in both retrieval directions at every evaluated missingness rate. However, Shot2Story and FLARE include conditions where refinement offers little benefit or underperforms the backbone.

On ActivityNet text-to-video retrieval, \why and \wrhg improve over \hypergram by $3.7$ and $4.1$ points in the complete T--VA setting, compared with $1.0$ point for \hy. Figure~\ref{fig:miss_t2v_hypergram} shows that both weighted configurations remain above the backbone at all four nonzero missingness rates, whereas \hy falls below it. The video-to-text results in Figure~\ref{fig:miss_v2t_hypergram} show the same preference for the weighted configurations: their initial gains of 11.9 and 12.1 points decrease with missingness but remain positive, while \hy falls below \hypergram at $90$\% missingness.

VATEX illustrates the importance of retrieval direction. All three configurations remain above \hypergram in text-to-video retrieval, although their relative ordering changes across the sweep. In video-to-text retrieval, \wrhg maintains a positive margin throughout, while \why's advantage approaches zero at $90$\% missingness. \hy falls below the backbone at $75$\% and $90$\% missingness. The relational configuration therefore provides the most persistent video-to-text advantage on this benchmark.

The Shot2Story results establish a limitation. In text-to-video retrieval, \wrhg's advantage becomes negligible at $90$\% missingness, while \hy and \why underperform the backbone at several rates. In video-to-text retrieval, all three configurations fall below \hypergram at the highest missingness rate. The improvement observed with complete observations therefore does not consistently transfer to severely incomplete clips on this benchmark.

FLARE also exhibits non-monotonic behavior. In text-to-video retrieval, all three configurations underperform \hypergram at $25$\%, $50$\%, and 75\% missingness, then show gains at $90$\%. In video-to-text retrieval, all three fall below the backbone at $75$\% missingness and recover positive gains at $90$\%. Since these curves report differences from a backbone whose performance also changes with missingness, such recoveries do not imply an increase in absolute recall.

Taken together, Figures~\ref{fig:miss_t2v_gram}--\ref{fig:miss_v2t_hypergram} show that the weighted configurations frequently preserve their advantage under incomplete observations, with particularly consistent results for \gram. The exceptions with \hypergram demonstrate that this advantage is not universal. Moreover, the hypergraph is absent during evaluation: the observed benefits arise through the learned encoder parameters, rather than through reconstruction of missing streams or information exchange between evaluation clips.

\section{Discussion}
\label{sec:Discussion}

\paragraph{Why training-time exchange helps.} The refinement never touches the test-time score, yet all three configurations improve both backbones on almost every video setting. The only path from the hypergraph to retrieval is the encoder weights, so the gains indicate that the encoders learn more useful per-modality embeddings when related clips can supply evidence during training. Several of the largest gains occur where the backbone is weak; for example, in ActivityNet video-to-text, \gram obtains $49.7$ R@1 and \wrhg $63.0$, and in FLARE in the T--VA setting, \wrhg raises \gram's text-to-video R@1 from $46.0$ to $53.9$. In this sense the hypergraph acts as a fusion mechanism that operates across samples during learning rather than across modalities at inference.

\paragraph{Weighting versus relational structure.} The two extensions address different weaknesses of \hy, and the results separate them only partly. The weighted configuration produces the clearest change: on ActivityNet it raises video-to-text R@1 by about six points over \hy with both backbones, and it keeps the gain under missing modalities where \hy falls below the backbone. Typed relations add a further but smaller improvement, mainly on Shot2Story and MSR-VTT, and on several benchmarks \why and \wrhg are within a few tenths of a point of each other. A plausible reading is that deciding \emph{how much} a view or a neighbour contributes matters more for low-quality data than deciding \emph{what kind} of link connects them. \hy remains the best configuration for MSVD text-to-video and for SoundDescs, so the added weighting and typing are not uniformly beneficial.
\paragraph{Asymmetry between retrieval directions.} Gains are generally larger for video-to-text than for text-to-video retrieval, reaching $13.3$ points on ActivityNet against $4.3$ in the other direction. The backbone scores are the same in both directions, but the rankings are formed over different sets. One possible explanation is that the refinement acts on the video embeddings, which are the side exposed to related clips during training, so a better-organised video space benefits most directly the direction in which videos act as queries.
\paragraph{Fusion under missing and unequal modalities.} The missing-modality study shows that the benefit of training-time fusion largely survives when views disappear at test time, although the models never see incomplete clips during training. With \gram, both weighted configurations remain above the backbone on all seven benchmarks at every rate, suggesting that learned vertex weights reduce the embeddings' dependence on any single stream. The \hypergram results are less regular, and on two benchmarks with long or multi-shot videos some conditions show no benefit, possibly because removing a stream there changes which modality carries the most information. On the audio benchmarks, gains are smaller and less consistent, which indicates that exchange between related clips is most useful when each clip offers several complementary streams to fuse.

\paragraph{Cost and applicability.} Since no part of the hypergraph survives training, the refined models have the same parameters at inference, the same encoding cost and the same scoring function as their backbones, unlike test-time re-ranking and query-bank methods~\cite{zhong2017reranking,bogolin2022qbnorm}. The approach can therefore be added to an existing volume-based retrieval system without changing its deployment; the only extra cost is a small hypergraph module during the same one-epoch training.
\section{Limitations}
\label{sec:Limitations}
\paragraph{Backbone coverage.} All compared models use the VAST encoders. We apply our models (\hy, \why, and \wrhg) to the two volume-based backbones, \gram and \hypergram. TRIANGLE~\cite{cicchetti2026triangle} is evaluated exclusively as a baseline because it is designed for exactly three modalities. Its score corresponds to the area spanned by three embeddings, the three-vector case of the \gram volume; it therefore cannot incorporate subtitles as a fourth stream, and its underlying principle is already represented by \gram. PMRL~\cite{liu2026principled} is also evaluated exclusively as a baseline because it adopts a different formulation: it aligns modalities by maximizing the leading singular value of their stacked embeddings, ranks candidates using a single fused representation, and assumes complete modality availability. In contrast, our models refine separate per-modality embeddings. Integrating them with PMRL would therefore require adapting its objective rather than directly incorporating our module. Extensions to PMRL and to encoders pretrained on other corpora remain directions for future research.
\paragraph{Training-time refinement only.} Our models employ the hypergraph feature fusion exclusively during training, with its influence incorporated into the encoder weights. At inference, each clip is encoded and scored independently using the original backbone. Although this design preserves the inference cost, it constrains the model's adaptability. Specifically, the deployed model cannot exploit gallery-specific structure, adapt to individual queries, or account for the particular pattern of missing streams encountered at test time. Such adaptation would require an explicit inference-time component, such as a lightweight graph constructed over the gallery, which lies beyond the scope of this work.
\paragraph{Training scale and single runs.} Following \gram, \pmrl, and \hypergram, we perform one epoch of continued pretraining on the 150K-clip subset of VAST-27M, of which 136{,}694 clips remained recoverable. Our conclusions are therefore restricted to this continued-pretraining regime and do not extend to training on the full corpus. Consistent with these studies, which do not report variability across repeated training runs, we train each configuration once using a fixed seed. All runs begin from the same pretrained checkpoint and differ only in the small hypergraph module. We therefore assess the consistency of improvements across eleven benchmarks and two backbones, rather than quantifying variability across repeated runs.

\paragraph{Simulated missing modalities.} None of the compared backbones evaluates retrieval with missing modalities. We therefore adopt the protocol introduced in our conference version~\cite{nag2027hyvol}, which, consistent with prior robustness studies~\cite{ma2022multimodal}, removes one randomly selected stream per clip at test time. This controlled protocol ensures identical evaluation conditions across methods. However, it only approximates the structured patterns of missing data encountered in real-world collections, in which entire sources or recording conditions may be absent.

\paragraph{Checkpoint selection.} Following the released training configurations of \gram, \tri, \pmrl, and \hypergram, we select the checkpoint by validating on MSR-VTT, which is also one of our evaluation benchmarks. We never use the remaining ten benchmarks during training or model selection, and the gains on them are similar to those on MSR-VTT, indicating that our conclusions do not depend on this choice. A held-out validation set drawn from the training distribution would make the evaluation fully independent of model selection.
\section{Conclusion}
\label{sec:conclusion}

We presented a hypergraph refinement framework for multimodal retrieval that models information exchange among observed modalities within a candidate and across semantically related training candidates. The framework includes \hy as a special case and accommodates two additional configurations: \why, which weights modality contributions and semantic connections, and \wrhg, which additionally distinguishes relation types and participant roles. Missing modalities are naturally excluded from the hypergraph without imputation. We restrict refinement to training, while inference retains the corresponding backbone's encoder outputs and retrieval scoring function.

Evaluation on $11$ zero-shot benchmarks demonstrates improvements across video--text and audio--text retrieval. With \gram, \wrhg improves over the backbone in all $17$ text-to-video settings and all $17$ video-to-text settings. With \hypergram, it improves $16$ of the $17$ text-to-video settings and all 17 video-to-text settings. The missing-modality evaluation on seven video benchmarks further shows that the weighted configurations often retain their advantage when individual clips lose an observed stream.

The results also identify the boundaries of these benefits. No refinement configuration is uniformly strongest across datasets and retrieval directions. \hy remains competitive or superior in several settings, audio-only retrieval exhibits smaller and less consistent gains, and some incomplete-modality conditions favor the unrefined \hypergram backbone. These findings support training-time hypergraph refinement as an effective approach to improving retrieval across evaluated settings, while showing that the value of weighting and relational modeling depends on the available information, backbone, and task.

\section*{CRediT authorship contribution statement}
\textbf{Anindya Nag:} Conceptualization, Investigation, Methodology, Writing - Original draft preparation.
\textbf{Ambuj Mehrish:} Methodology, Validation, Writing - Original draft preparation, Writing - Review \& Editing.
\textbf{Sebastiano Vascon:} Supervision, Methodology, Conceptualization, Writing - Review \& Editing.

\section*{Funding}
This work was supported by the DoE 2023-2027 (MUR, AIS.DIP.ECCELLENZA2023\_27.FF project).

\section*{Acknowledgements}
This work was supported by the European Union’s Horizon Europe research and innovation programme under the Marie Skłodowska-Curie grant agreement No. 101205348 (CASPER). This work was supported by the DoE 2023-2027 (MUR, AIS.DIP.ECCELLENZA2023\_27.FF project). We acknowledge the EuroHPC Joint Undertaking for awarding this project access to the EuroHPC supercomputer LEONARDO, hosted by CINECA (Italy) and the LEONARDO consortium, through the EuroHPC AI Factories “AI for Science and Collaborative EU Projects” Access call (proposal No. EHPC-AIF-2026SC01-041). We further acknowledge the CINECA award under the ISCRA initiative (Class C projects IsCd5-CASPER-A and IsCd5\_GMEG), for the availability of high-performance computing resources and support.

\section*{Declaration of competing interest}
The authors declare that they have no known competing financial interests or personal relationships that could have appeared to influence the work reported in this paper.

\section*{Data availability}
The code and trained weights of \textsc{HyVol}, our conference model, are available \href{https://github.com/CVML-CFU/HyVol}{here}. The trained weights of \textsc{wHyVol} and \textsc{wRHG} for both backbones are available \href{https://huggingface.co/CVML-CFU/HyVol-PlusPlus}{here}, and their code will be released upon publication. All datasets used in this work are publicly available from their original sources.

\bibliographystyle{elsarticle-num-names}
\bibliography{main}

\clearpage
\appendix

\section{Use of Large Language Models}
\label{sec:supp-llm}
We used large language models only to assist with the writing, namely to correct grammar, refine wording and readability, and shorten the text where needed. The authors originated the research idea, method design, experimental protocol, analyses, and all reported results. The authors checked every suggested change against their experimental records and accept full responsibility for the content of this paper.

\section{Reproducibility}
\label{sec:supp-repro}
We have made the code and trained weights of \textsc{HyVol} publicly available \href{https://github.com/CVML-CFU/HyVol}{here}. The trained weights of \textsc{wHyVol} and \textsc{wRHG} for both backbones are available \href{https://huggingface.co/CVML-CFU/HyVol-PlusPlus}{here}, and we will release the corresponding training and evaluation code, configuration files, and the scripts that produce our tables and figures upon publication. We use one configuration per variant for all eleven benchmarks (eight text--video and three text--audio retrieval tasks) and do not tune the learning rate, neighbourhood size, or schedule for any individual benchmark. We train all models on the same 150K-clip subset of VAST-27M, of which we could download 136{,}694 clips with both video and audio.

For \textsc{Gram}, \textsc{Triangle} and \textsc{PMRL}, we use the checkpoints made available by their authors. Since no checkpoint is available for \textsc{HyperGram}, we train it ourselves with the authors' public code and their published training recipe. Every evaluated method is initialised from the VAST foundation model and tested on identical test data. Published numbers from other works are reported only as reference values.

\section{Implementation details}
\label{app:hparams}
Table~\ref{tab:hyper_parameters} lists the complete configuration of \hy, \why, and \wrhg, complementing the summary in Section~\ref{sec:model_configuration}. The three configurations share the encoders, optimisation schedule, and data settings, and differ only in how the hypergraph is weighted and propagated: \why and \wrhg learn vertex and pair weights, and \wrhg additionally uses typed relational messages. The same values are used for both backbones and all eleven benchmarks without benchmark-specific tuning, and the hypergraph is used only during training.
\begin{table}[t]
\centering
\caption{\textbf{Hyperparameters.} One configuration is used for both backbones and all eleven benchmarks, without benchmark-specific tuning; settings that differ between configurations name the configurations they apply to.}
\label{tab:hyper_parameters}
\scriptsize
\setlength{\tabcolsep}{3pt}
\renewcommand{\arraystretch}{0.95}
\begin{tabular}{@{}l>{\raggedright\arraybackslash}p{0.52\columnwidth}@{}}
\toprule
\multicolumn{2}{@{}l}{\emph{Backbone}} \\
Initialization & VAST foundation checkpoint \\
Vision encoder & EVA-CLIP ViT-g/14 \\
Audio encoder & BEATs \\
Text encoder & BERT-base (caption and subtitle) \\
Projection dimension $d$ & 512, followed by $\ell_2$ normalisation \\
Trainable modules & all encoders, projections, hypergraph, and heads \\
\midrule
\multicolumn{2}{@{}l}{\emph{Hypergraph construction}} \\
Neighbours $k$ / adaptive cap & 12 / $\lfloor B_{\mathrm{shard}}/4\rfloor=16$ \\
Neighbourhood & mutual top-$k$ cosine on stop-gradient captions \\
Similarity filter & disabled \\
Semantic-edge dropout & 0.3 \\
Relation families (\wrhg) & document, same-modality, cross-modality, joint-candidate \\
\midrule
\multicolumn{2}{@{}l}{\emph{Learned weights}} \\
Semantic weights & softmax attention (\hy); kernel of Eq.~\eqref{eq:weights} (\why, \wrhg) \\
Affinity temperature $\tau_{\mathrm{a}}$ & 0.5, learned (\why, \wrhg) \\
Affinity projection $W_{\mathrm{a}}$ & identity initialisation (\why, \wrhg) \\
Vertex weight $\beta_i^m$ & fixed at 1 (\hy); learned in $(0,2)$, initialised to 1 (\why, \wrhg) \\
\midrule
\multicolumn{2}{@{}l}{\emph{Propagation}} \\
Operator & additive (\hy, \why); relational (\wrhg) \\
Layers $L$ & 2 \\
Activation & GELU (\hy, \why); ReLU (\wrhg); identity in the final layer \\
Mixing $\alpha_\ell$, Eq.~\eqref{eq:relational} (\wrhg) & initialised to 0.5, learned \\
\midrule
\multicolumn{2}{@{}l}{\emph{Losses}} \\
Retrieval objective & backbone volume loss, symmetric cross-entropy \\
Backbone temperature & loaded from the checkpoint, learned \\
Label smoothing & 0.1 (volume-based losses) \\
Alignment weight $\lambda_{\mathrm{doc}}$ & 1.0 \\
Variance weight $\lambda_{\mathrm{reg}}$ / $\epsilon$ & 0.1 / $10^{-4}$ \\
Matching weight $\lambda_{\mathrm{itm}}$ & 0.1 \\
\midrule
\multicolumn{2}{@{}l}{\emph{\hypergram{} backbone}} \\
Curvature & initialised to 1, learned \\
Euclidean / hyperbolic weights & initialised to 0.5 / 0.5, learned \\
Hyperbolic gradient clip & 1.0 \\
\midrule
\multicolumn{2}{@{}l}{\emph{Optimisation}} \\
Optimiser & AdamW, $\beta=(0.9,0.98)$ \\
Learning rate & $2\times10^{-5}$ backbone; $5\times10^{-7}$ visual encoder; $5\times10^{-4}$ hypergraph \\
Weight decay / gradient clip & 0.01 / 2.0 \\
Schedule & linear; warm-up ratio 0.1 \\
Batch size & 256 global; 64 per GPU \\
Epochs / steps & 1 / 533 \\
Training frames / audio & 2 / one 10\,s segment \\
Precision / parallelism & fp16 / DDP over four A100 GPUs \\
Seed & 50 (single run) \\
\midrule
\multicolumn{2}{@{}l}{\emph{Model selection and inference}} \\
Validation & MSR-VTT 1k, T--VAS, ten times per epoch \\
Selection criterion & text-to-video R@1 of the matching score \\
Sampled frames & 8 \\
Audio segments & 1 or 2 of 10\,s, per benchmark \\
Maximum query length & 40 or 70 tokens, per benchmark \\
Hypergraph at test time & removed; encoder outputs scored directly \\
Test-time missingness rates $r$ & 0, 25, 50, 75, 90\% \\
\bottomrule
\end{tabular}
\end{table}

\section{Additional Results and Analysis}
\label{app:additional}

Table~\ref{tab:miss_all} reports text-to-video and video-to-text R@1 on seven video benchmarks as the share of gallery clips missing one modality grows from 0\% to 90\%. Recall falls for every method as clips become incomplete, yet the refined models generally remain clearly ahead of their backbones. On VALOR-32K video-to-text retrieval with \textsc{Gram}, for example, \textsc{wRHG} leads by 9.1, 8.8, and 7.0 points at 25\%, 50\%, and 75\% missingness, and its advantage persists at 6.6 points even when 90\% of the clips lack one modality.

\begin{table*}[t]
\centering
\caption{Retrieval under missing modalities (R@1) on the seven video benchmarks. A fraction $r$ of the gallery clips loses one randomly chosen modality; the per-clip masks are fixed and identical across methods, and $r{=}0$ reproduces the main protocol. MSR-VTT and VATEX use the T--VAS modality set, and the other benchmarks use T--VA. Numbers in parentheses give the change with respect to the backbone (\textsc{Gram} or \textsc{HyperGram}); green marks an improvement, red a drop, and gray no change.}
\label{tab:miss_all}
\footnotesize
\setlength{\tabcolsep}{2pt}

\renewcommand{\arraystretch}{0.95}
\resizebox{0.95\textwidth}{!}{%

\begin{tabular}{@{}llccccccccccc@{}}
\toprule
 & & \multicolumn{5}{c}{T$\to$V} & & \multicolumn{5}{c}{V$\to$T} \\
\cmidrule(lr){3-7}\cmidrule(lr){9-13}
Dataset & Method & 0\% & 25\% & 50\% & 75\% & 90\% & & 0\% & 25\% & 50\% & 75\% & 90\% \\
\midrule
MSR-VTT~\cite{MSRVTT} & \textsc{Gram}$^\star$ & 51.8 & 49.2 & 44.0 & 41.4 & 39.5 &  & 50.1 & 46.5 & 42.4 & 38.1 & 36.4 \\
(T--VAS) & \quad +\textsc{HyVol} & 55.8\,{\scriptsize\textcolor{green!50!black}{(+4.0)}} & 52.7\,{\scriptsize\textcolor{green!50!black}{(+3.5)}} & 47.2\,{\scriptsize\textcolor{green!50!black}{(+3.2)}} & 41.6\,{\scriptsize\textcolor{green!50!black}{(+0.2)}} & 39.3\,{\scriptsize\textcolor{red!70!black}{($-$0.2)}} &  & 58.1\,{\scriptsize\textcolor{green!50!black}{(+8.0)}} & 54.1\,{\scriptsize\textcolor{green!50!black}{(+7.6)}} & 48.0\,{\scriptsize\textcolor{green!50!black}{(+5.6)}} & 42.6\,{\scriptsize\textcolor{green!50!black}{(+4.5)}} & 39.4\,{\scriptsize\textcolor{green!50!black}{(+3.0)}} \\
 & \quad +\textsc{wHyVol} & 56.8\,{\scriptsize\textcolor{green!50!black}{(+5.0)}} & 53.7\,{\scriptsize\textcolor{green!50!black}{(+4.5)}} & 48.0\,{\scriptsize\textcolor{green!50!black}{(+4.0)}} & 43.9\,{\scriptsize\textcolor{green!50!black}{(+2.5)}} & 41.2\,{\scriptsize\textcolor{green!50!black}{(+1.7)}} &  & 58.0\,{\scriptsize\textcolor{green!50!black}{(+7.9)}} & 53.6\,{\scriptsize\textcolor{green!50!black}{(+7.1)}} & 47.9\,{\scriptsize\textcolor{green!50!black}{(+5.5)}} & 43.1\,{\scriptsize\textcolor{green!50!black}{(+5.0)}} & 40.7\,{\scriptsize\textcolor{green!50!black}{(+4.3)}} \\
 & \quad +\textsc{wRHG} & 57.4\,{\scriptsize\textcolor{green!50!black}{(+5.6)}} & 53.5\,{\scriptsize\textcolor{green!50!black}{(+4.3)}} & 48.5\,{\scriptsize\textcolor{green!50!black}{(+4.5)}} & 44.4\,{\scriptsize\textcolor{green!50!black}{(+3.0)}} & 41.3\,{\scriptsize\textcolor{green!50!black}{(+1.8)}} &  & 58.0\,{\scriptsize\textcolor{green!50!black}{(+7.9)}} & 53.6\,{\scriptsize\textcolor{green!50!black}{(+7.1)}} & 48.6\,{\scriptsize\textcolor{green!50!black}{(+6.2)}} & 43.1\,{\scriptsize\textcolor{green!50!black}{(+5.0)}} & 40.2\,{\scriptsize\textcolor{green!50!black}{(+3.8)}} \\
\cmidrule(l){2-13}
 & \textsc{HyperGram}$^\dag$ & 52.3 & 50.2 & 45.4 & 41.2 & 39.2 &  & 50.9 & 47.6 & 43.3 & 39.1 & 36.7 \\
 & \quad +\textsc{HyVol} & 56.1\,{\scriptsize\textcolor{green!50!black}{(+3.8)}} & 52.3\,{\scriptsize\textcolor{green!50!black}{(+2.1)}} & 47.2\,{\scriptsize\textcolor{green!50!black}{(+1.8)}} & 42.3\,{\scriptsize\textcolor{green!50!black}{(+1.1)}} & 39.9\,{\scriptsize\textcolor{green!50!black}{(+0.7)}} &  & 59.2\,{\scriptsize\textcolor{green!50!black}{(+8.3)}} & 54.7\,{\scriptsize\textcolor{green!50!black}{(+7.1)}} & 47.9\,{\scriptsize\textcolor{green!50!black}{(+4.6)}} & 42.6\,{\scriptsize\textcolor{green!50!black}{(+3.5)}} & 39.5\,{\scriptsize\textcolor{green!50!black}{(+2.8)}} \\
 & \quad +\textsc{wHyVol} & 57.0\,{\scriptsize\textcolor{green!50!black}{(+4.7)}} & 53.2\,{\scriptsize\textcolor{green!50!black}{(+3.0)}} & 48.9\,{\scriptsize\textcolor{green!50!black}{(+3.5)}} & 43.3\,{\scriptsize\textcolor{green!50!black}{(+2.1)}} & 40.8\,{\scriptsize\textcolor{green!50!black}{(+1.6)}} &  & 58.8\,{\scriptsize\textcolor{green!50!black}{(+7.9)}} & 54.3\,{\scriptsize\textcolor{green!50!black}{(+6.7)}} & 47.9\,{\scriptsize\textcolor{green!50!black}{(+4.6)}} & 42.5\,{\scriptsize\textcolor{green!50!black}{(+3.4)}} & 40.2\,{\scriptsize\textcolor{green!50!black}{(+3.5)}} \\
 & \quad +\textsc{wRHG} & 57.3\,{\scriptsize\textcolor{green!50!black}{(+5.0)}} & 53.7\,{\scriptsize\textcolor{green!50!black}{(+3.5)}} & 48.9\,{\scriptsize\textcolor{green!50!black}{(+3.5)}} & 43.8\,{\scriptsize\textcolor{green!50!black}{(+2.6)}} & 41.6\,{\scriptsize\textcolor{green!50!black}{(+2.4)}} &  & 59.3\,{\scriptsize\textcolor{green!50!black}{(+8.4)}} & 54.3\,{\scriptsize\textcolor{green!50!black}{(+6.7)}} & 48.3\,{\scriptsize\textcolor{green!50!black}{(+5.0)}} & 43.5\,{\scriptsize\textcolor{green!50!black}{(+4.4)}} & 40.9\,{\scriptsize\textcolor{green!50!black}{(+4.2)}} \\
\midrule
DiDeMo~\cite{DIDEMO} & \textsc{Gram}$^\star$ & 50.6 & 45.5 & 40.6 & 36.6 & 33.5 &  & 49.0 & 42.5 & 36.7 & 32.1 & 28.6 \\
(T--VA) & \quad +\textsc{HyVol} & 54.6\,{\scriptsize\textcolor{green!50!black}{(+4.0)}} & 47.5\,{\scriptsize\textcolor{green!50!black}{(+2.0)}} & 42.4\,{\scriptsize\textcolor{green!50!black}{(+1.8)}} & 38.1\,{\scriptsize\textcolor{green!50!black}{(+1.5)}} & 35.2\,{\scriptsize\textcolor{green!50!black}{(+1.7)}} &  & 54.1\,{\scriptsize\textcolor{green!50!black}{(+5.1)}} & 47.5\,{\scriptsize\textcolor{green!50!black}{(+5.0)}} & 41.1\,{\scriptsize\textcolor{green!50!black}{(+4.4)}} & 36.1\,{\scriptsize\textcolor{green!50!black}{(+4.0)}} & 31.2\,{\scriptsize\textcolor{green!50!black}{(+2.6)}} \\
 & \quad +\textsc{wHyVol} & 53.8\,{\scriptsize\textcolor{green!50!black}{(+3.2)}} & 47.2\,{\scriptsize\textcolor{green!50!black}{(+1.7)}} & 42.0\,{\scriptsize\textcolor{green!50!black}{(+1.4)}} & 38.0\,{\scriptsize\textcolor{green!50!black}{(+1.4)}} & 34.0\,{\scriptsize\textcolor{green!50!black}{(+0.5)}} &  & 55.5\,{\scriptsize\textcolor{green!50!black}{(+6.5)}} & 47.1\,{\scriptsize\textcolor{green!50!black}{(+4.6)}} & 40.5\,{\scriptsize\textcolor{green!50!black}{(+3.8)}} & 35.1\,{\scriptsize\textcolor{green!50!black}{(+3.0)}} & 31.3\,{\scriptsize\textcolor{green!50!black}{(+2.7)}} \\
 & \quad +\textsc{wRHG} & 54.1\,{\scriptsize\textcolor{green!50!black}{(+3.5)}} & 46.9\,{\scriptsize\textcolor{green!50!black}{(+1.4)}} & 41.5\,{\scriptsize\textcolor{green!50!black}{(+0.9)}} & 36.9\,{\scriptsize\textcolor{green!50!black}{(+0.3)}} & 33.8\,{\scriptsize\textcolor{green!50!black}{(+0.3)}} &  & 55.0\,{\scriptsize\textcolor{green!50!black}{(+6.0)}} & 47.4\,{\scriptsize\textcolor{green!50!black}{(+4.9)}} & 40.8\,{\scriptsize\textcolor{green!50!black}{(+4.1)}} & 35.9\,{\scriptsize\textcolor{green!50!black}{(+3.8)}} & 31.8\,{\scriptsize\textcolor{green!50!black}{(+3.2)}} \\
\cmidrule(l){2-13}
 & \textsc{HyperGram}$^\dag$ & 50.4 & 45.0 & 39.9 & 36.6 & 32.8 &  & 50.8 & 43.7 & 37.6 & 33.6 & 29.9 \\
 & \quad +\textsc{HyVol} & 54.2\,{\scriptsize\textcolor{green!50!black}{(+3.8)}} & 47.8\,{\scriptsize\textcolor{green!50!black}{(+2.8)}} & 41.9\,{\scriptsize\textcolor{green!50!black}{(+2.0)}} & 38.8\,{\scriptsize\textcolor{green!50!black}{(+2.2)}} & 35.1\,{\scriptsize\textcolor{green!50!black}{(+2.3)}} &  & 54.7\,{\scriptsize\textcolor{green!50!black}{(+3.9)}} & 48.2\,{\scriptsize\textcolor{green!50!black}{(+4.5)}} & 40.5\,{\scriptsize\textcolor{green!50!black}{(+2.9)}} & 35.6\,{\scriptsize\textcolor{green!50!black}{(+2.0)}} & 31.0\,{\scriptsize\textcolor{green!50!black}{(+1.1)}} \\
 & \quad +\textsc{wHyVol} & 54.6\,{\scriptsize\textcolor{green!50!black}{(+4.2)}} & 48.1\,{\scriptsize\textcolor{green!50!black}{(+3.1)}} & 42.5\,{\scriptsize\textcolor{green!50!black}{(+2.6)}} & 37.7\,{\scriptsize\textcolor{green!50!black}{(+1.1)}} & 34.3\,{\scriptsize\textcolor{green!50!black}{(+1.5)}} &  & 55.1\,{\scriptsize\textcolor{green!50!black}{(+4.3)}} & 46.8\,{\scriptsize\textcolor{green!50!black}{(+3.1)}} & 39.9\,{\scriptsize\textcolor{green!50!black}{(+2.3)}} & 34.6\,{\scriptsize\textcolor{green!50!black}{(+1.0)}} & 30.5\,{\scriptsize\textcolor{green!50!black}{(+0.6)}} \\
 & \quad +\textsc{wRHG} & 54.6\,{\scriptsize\textcolor{green!50!black}{(+4.2)}} & 48.1\,{\scriptsize\textcolor{green!50!black}{(+3.1)}} & 42.5\,{\scriptsize\textcolor{green!50!black}{(+2.6)}} & 38.6\,{\scriptsize\textcolor{green!50!black}{(+2.0)}} & 34.5\,{\scriptsize\textcolor{green!50!black}{(+1.7)}} &  & 55.5\,{\scriptsize\textcolor{green!50!black}{(+4.7)}} & 47.2\,{\scriptsize\textcolor{green!50!black}{(+3.5)}} & 40.8\,{\scriptsize\textcolor{green!50!black}{(+3.2)}} & 34.7\,{\scriptsize\textcolor{green!50!black}{(+1.1)}} & 31.0\,{\scriptsize\textcolor{green!50!black}{(+1.1)}} \\
\midrule
ActivityNet~\cite{ACTIVITYNET} & \textsc{Gram}$^\star$ & 56.4 & 50.4 & 44.4 & 38.1 & 34.2 &  & 49.7 & 43.9 & 38.4 & 31.7 & 28.2 \\
(T--VA) & \quad +\textsc{HyVol} & 56.9\,{\scriptsize\textcolor{green!50!black}{(+0.5)}} & 49.0\,{\scriptsize\textcolor{red!70!black}{($-$1.4)}} & 43.5\,{\scriptsize\textcolor{red!70!black}{($-$0.9)}} & 37.8\,{\scriptsize\textcolor{red!70!black}{($-$0.3)}} & 34.2\,{\scriptsize\textcolor{gray}{(0.0)}} &  & 56.6\,{\scriptsize\textcolor{green!50!black}{(+6.9)}} & 48.5\,{\scriptsize\textcolor{green!50!black}{(+4.6)}} & 41.6\,{\scriptsize\textcolor{green!50!black}{(+3.2)}} & 32.9\,{\scriptsize\textcolor{green!50!black}{(+1.2)}} & 29.0\,{\scriptsize\textcolor{green!50!black}{(+0.8)}} \\
 & \quad +\textsc{wHyVol} & 60.8\,{\scriptsize\textcolor{green!50!black}{(+4.4)}} & 54.3\,{\scriptsize\textcolor{green!50!black}{(+3.9)}} & 47.7\,{\scriptsize\textcolor{green!50!black}{(+3.3)}} & 40.9\,{\scriptsize\textcolor{green!50!black}{(+2.8)}} & 36.5\,{\scriptsize\textcolor{green!50!black}{(+2.3)}} &  & 62.7\,{\scriptsize\textcolor{green!50!black}{(+13.0)}} & 53.8\,{\scriptsize\textcolor{green!50!black}{(+9.9)}} & 44.9\,{\scriptsize\textcolor{green!50!black}{(+6.5)}} & 36.3\,{\scriptsize\textcolor{green!50!black}{(+4.6)}} & 31.2\,{\scriptsize\textcolor{green!50!black}{(+3.0)}} \\
 & \quad +\textsc{wRHG} & 60.7\,{\scriptsize\textcolor{green!50!black}{(+4.3)}} & 54.1\,{\scriptsize\textcolor{green!50!black}{(+3.7)}} & 47.9\,{\scriptsize\textcolor{green!50!black}{(+3.5)}} & 40.7\,{\scriptsize\textcolor{green!50!black}{(+2.6)}} & 36.6\,{\scriptsize\textcolor{green!50!black}{(+2.4)}} &  & 63.0\,{\scriptsize\textcolor{green!50!black}{(+13.3)}} & 54.2\,{\scriptsize\textcolor{green!50!black}{(+10.3)}} & 45.4\,{\scriptsize\textcolor{green!50!black}{(+7.0)}} & 36.7\,{\scriptsize\textcolor{green!50!black}{(+5.0)}} & 31.7\,{\scriptsize\textcolor{green!50!black}{(+3.5)}} \\
\cmidrule(l){2-13}
 & \textsc{HyperGram}$^\dag$ & 56.8 & 50.9 & 45.2 & 38.9 & 35.2 &  & 50.9 & 44.6 & 39.1 & 32.6 & 28.9 \\
 & \quad +\textsc{HyVol} & 57.8\,{\scriptsize\textcolor{green!50!black}{(+1.0)}} & 48.7\,{\scriptsize\textcolor{red!70!black}{($-$2.2)}} & 43.7\,{\scriptsize\textcolor{red!70!black}{($-$1.5)}} & 38.0\,{\scriptsize\textcolor{red!70!black}{($-$0.9)}} & 34.1\,{\scriptsize\textcolor{red!70!black}{($-$1.1)}} &  & 56.4\,{\scriptsize\textcolor{green!50!black}{(+5.5)}} & 48.7\,{\scriptsize\textcolor{green!50!black}{(+4.1)}} & 41.3\,{\scriptsize\textcolor{green!50!black}{(+2.2)}} & 32.6\,{\scriptsize\textcolor{gray}{(0.0)}} & 28.3\,{\scriptsize\textcolor{red!70!black}{($-$0.6)}} \\
 & \quad +\textsc{wHyVol} & 60.5\,{\scriptsize\textcolor{green!50!black}{(+3.7)}} & 54.2\,{\scriptsize\textcolor{green!50!black}{(+3.3)}} & 48.0\,{\scriptsize\textcolor{green!50!black}{(+2.8)}} & 40.9\,{\scriptsize\textcolor{green!50!black}{(+2.0)}} & 36.5\,{\scriptsize\textcolor{green!50!black}{(+1.3)}} &  & 62.8\,{\scriptsize\textcolor{green!50!black}{(+11.9)}} & 53.7\,{\scriptsize\textcolor{green!50!black}{(+9.1)}} & 45.0\,{\scriptsize\textcolor{green!50!black}{(+5.9)}} & 36.6\,{\scriptsize\textcolor{green!50!black}{(+4.0)}} & 31.4\,{\scriptsize\textcolor{green!50!black}{(+2.5)}} \\
 & \quad +\textsc{wRHG} & 60.9\,{\scriptsize\textcolor{green!50!black}{(+4.1)}} & 54.5\,{\scriptsize\textcolor{green!50!black}{(+3.6)}} & 48.3\,{\scriptsize\textcolor{green!50!black}{(+3.1)}} & 41.1\,{\scriptsize\textcolor{green!50!black}{(+2.2)}} & 36.8\,{\scriptsize\textcolor{green!50!black}{(+1.6)}} &  & 63.0\,{\scriptsize\textcolor{green!50!black}{(+12.1)}} & 53.9\,{\scriptsize\textcolor{green!50!black}{(+9.3)}} & 45.0\,{\scriptsize\textcolor{green!50!black}{(+5.9)}} & 36.4\,{\scriptsize\textcolor{green!50!black}{(+3.8)}} & 31.3\,{\scriptsize\textcolor{green!50!black}{(+2.4)}} \\
\midrule
VATEX~\cite{wang2019vatex} & \textsc{Gram}$^\star$ & 88.4 & 80.7 & 74.5 & 67.7 & 64.5 &  & 86.5 & 79.1 & 73.5 & 67.3 & 64.0 \\
(T--VAS) & \quad +\textsc{HyVol} & 94.0\,{\scriptsize\textcolor{green!50!black}{(+5.6)}} & 85.8\,{\scriptsize\textcolor{green!50!black}{(+5.1)}} & 78.0\,{\scriptsize\textcolor{green!50!black}{(+3.5)}} & 70.8\,{\scriptsize\textcolor{green!50!black}{(+3.1)}} & 66.4\,{\scriptsize\textcolor{green!50!black}{(+1.9)}} &  & 93.8\,{\scriptsize\textcolor{green!50!black}{(+7.3)}} & 84.9\,{\scriptsize\textcolor{green!50!black}{(+5.8)}} & 77.3\,{\scriptsize\textcolor{green!50!black}{(+3.8)}} & 69.8\,{\scriptsize\textcolor{green!50!black}{(+2.5)}} & 65.2\,{\scriptsize\textcolor{green!50!black}{(+1.2)}} \\
 & \quad +\textsc{wHyVol} & 93.5\,{\scriptsize\textcolor{green!50!black}{(+5.1)}} & 84.9\,{\scriptsize\textcolor{green!50!black}{(+4.2)}} & 77.0\,{\scriptsize\textcolor{green!50!black}{(+2.5)}} & 70.5\,{\scriptsize\textcolor{green!50!black}{(+2.8)}} & 67.5\,{\scriptsize\textcolor{green!50!black}{(+3.0)}} &  & 94.4\,{\scriptsize\textcolor{green!50!black}{(+7.9)}} & 86.1\,{\scriptsize\textcolor{green!50!black}{(+7.0)}} & 78.4\,{\scriptsize\textcolor{green!50!black}{(+4.9)}} & 71.5\,{\scriptsize\textcolor{green!50!black}{(+4.2)}} & 67.5\,{\scriptsize\textcolor{green!50!black}{(+3.5)}} \\
 & \quad +\textsc{wRHG} & 94.0\,{\scriptsize\textcolor{green!50!black}{(+5.6)}} & 85.6\,{\scriptsize\textcolor{green!50!black}{(+4.9)}} & 77.3\,{\scriptsize\textcolor{green!50!black}{(+2.8)}} & 71.0\,{\scriptsize\textcolor{green!50!black}{(+3.3)}} & 68.4\,{\scriptsize\textcolor{green!50!black}{(+3.9)}} &  & 94.4\,{\scriptsize\textcolor{green!50!black}{(+7.9)}} & 86.1\,{\scriptsize\textcolor{green!50!black}{(+7.0)}} & 78.4\,{\scriptsize\textcolor{green!50!black}{(+4.9)}} & 71.2\,{\scriptsize\textcolor{green!50!black}{(+3.9)}} & 67.5\,{\scriptsize\textcolor{green!50!black}{(+3.5)}} \\
\cmidrule(l){2-13}
 & \textsc{HyperGram}$^\dag$ & 89.8 & 81.7 & 74.7 & 67.3 & 63.8 &  & 87.8 & 80.5 & 74.7 & 69.4 & 65.9 \\
 & \quad +\textsc{HyVol} & 93.3\,{\scriptsize\textcolor{green!50!black}{(+3.5)}} & 85.4\,{\scriptsize\textcolor{green!50!black}{(+3.7)}} & 78.2\,{\scriptsize\textcolor{green!50!black}{(+3.5)}} & 70.8\,{\scriptsize\textcolor{green!50!black}{(+3.5)}} & 66.8\,{\scriptsize\textcolor{green!50!black}{(+3.0)}} &  & 93.9\,{\scriptsize\textcolor{green!50!black}{(+6.1)}} & 83.5\,{\scriptsize\textcolor{green!50!black}{(+3.0)}} & 74.9\,{\scriptsize\textcolor{green!50!black}{(+0.2)}} & 68.2\,{\scriptsize\textcolor{red!70!black}{($-$1.2)}} & 63.3\,{\scriptsize\textcolor{red!70!black}{($-$2.6)}} \\
 & \quad +\textsc{wHyVol} & 93.5\,{\scriptsize\textcolor{green!50!black}{(+3.7)}} & 85.8\,{\scriptsize\textcolor{green!50!black}{(+4.1)}} & 77.3\,{\scriptsize\textcolor{green!50!black}{(+2.6)}} & 68.9\,{\scriptsize\textcolor{green!50!black}{(+1.6)}} & 65.2\,{\scriptsize\textcolor{green!50!black}{(+1.4)}} &  & 94.0\,{\scriptsize\textcolor{green!50!black}{(+6.2)}} & 84.7\,{\scriptsize\textcolor{green!50!black}{(+4.2)}} & 76.8\,{\scriptsize\textcolor{green!50!black}{(+2.1)}} & 70.8\,{\scriptsize\textcolor{green!50!black}{(+1.4)}} & 65.9\,{\scriptsize\textcolor{gray}{(0.0)}} \\
 & \quad +\textsc{wRHG} & 93.5\,{\scriptsize\textcolor{green!50!black}{(+3.7)}} & 85.2\,{\scriptsize\textcolor{green!50!black}{(+3.5)}} & 77.0\,{\scriptsize\textcolor{green!50!black}{(+2.3)}} & 70.5\,{\scriptsize\textcolor{green!50!black}{(+3.2)}} & 67.5\,{\scriptsize\textcolor{green!50!black}{(+3.7)}} &  & 94.0\,{\scriptsize\textcolor{green!50!black}{(+6.2)}} & 85.4\,{\scriptsize\textcolor{green!50!black}{(+4.9)}} & 77.3\,{\scriptsize\textcolor{green!50!black}{(+2.6)}} & 71.7\,{\scriptsize\textcolor{green!50!black}{(+2.3)}} & 68.2\,{\scriptsize\textcolor{green!50!black}{(+2.3)}} \\
\midrule
Shot2Story~\cite{han2023shot2story} & \textsc{Gram}$^\star$ & 70.5 & 62.4 & 54.0 & 46.3 & 42.2 &  & 66.2 & 58.2 & 50.7 & 42.9 & 37.9 \\
(T--VA) & \quad +\textsc{HyVol} & 74.8\,{\scriptsize\textcolor{green!50!black}{(+4.3)}} & 66.4\,{\scriptsize\textcolor{green!50!black}{(+4.0)}} & 58.5\,{\scriptsize\textcolor{green!50!black}{(+4.5)}} & 50.2\,{\scriptsize\textcolor{green!50!black}{(+3.9)}} & 45.2\,{\scriptsize\textcolor{green!50!black}{(+3.0)}} &  & 77.6\,{\scriptsize\textcolor{green!50!black}{(+11.4)}} & 67.1\,{\scriptsize\textcolor{green!50!black}{(+8.9)}} & 56.9\,{\scriptsize\textcolor{green!50!black}{(+6.2)}} & 47.6\,{\scriptsize\textcolor{green!50!black}{(+4.7)}} & 41.9\,{\scriptsize\textcolor{green!50!black}{(+4.0)}} \\
 & \quad +\textsc{wHyVol} & 76.3\,{\scriptsize\textcolor{green!50!black}{(+5.8)}} & 66.7\,{\scriptsize\textcolor{green!50!black}{(+4.3)}} & 57.5\,{\scriptsize\textcolor{green!50!black}{(+3.5)}} & 49.2\,{\scriptsize\textcolor{green!50!black}{(+2.9)}} & 44.4\,{\scriptsize\textcolor{green!50!black}{(+2.2)}} &  & 75.7\,{\scriptsize\textcolor{green!50!black}{(+9.5)}} & 66.2\,{\scriptsize\textcolor{green!50!black}{(+8.0)}} & 56.3\,{\scriptsize\textcolor{green!50!black}{(+5.6)}} & 47.1\,{\scriptsize\textcolor{green!50!black}{(+4.2)}} & 41.5\,{\scriptsize\textcolor{green!50!black}{(+3.6)}} \\
 & \quad +\textsc{wRHG} & 77.4\,{\scriptsize\textcolor{green!50!black}{(+6.9)}} & 68.3\,{\scriptsize\textcolor{green!50!black}{(+5.9)}} & 59.0\,{\scriptsize\textcolor{green!50!black}{(+5.0)}} & 50.5\,{\scriptsize\textcolor{green!50!black}{(+4.2)}} & 45.0\,{\scriptsize\textcolor{green!50!black}{(+2.8)}} &  & 76.3\,{\scriptsize\textcolor{green!50!black}{(+10.1)}} & 66.8\,{\scriptsize\textcolor{green!50!black}{(+8.6)}} & 56.7\,{\scriptsize\textcolor{green!50!black}{(+6.0)}} & 47.8\,{\scriptsize\textcolor{green!50!black}{(+4.9)}} & 42.0\,{\scriptsize\textcolor{green!50!black}{(+4.1)}} \\
\cmidrule(l){2-13}
 & \textsc{HyperGram}$^\dag$ & 75.0 & 67.4 & 58.5 & 50.6 & 45.2 &  & 73.1 & 65.6 & 56.4 & 47.7 & 42.3 \\
 & \quad +\textsc{HyVol} & 74.6\,{\scriptsize\textcolor{red!70!black}{($-$0.4)}} & 66.5\,{\scriptsize\textcolor{red!70!black}{($-$0.9)}} & 58.4\,{\scriptsize\textcolor{red!70!black}{($-$0.1)}} & 50.4\,{\scriptsize\textcolor{red!70!black}{($-$0.2)}} & 44.8\,{\scriptsize\textcolor{red!70!black}{($-$0.4)}} &  & 78.1\,{\scriptsize\textcolor{green!50!black}{(+5.0)}} & 67.4\,{\scriptsize\textcolor{green!50!black}{(+1.8)}} & 57.1\,{\scriptsize\textcolor{green!50!black}{(+0.7)}} & 48.0\,{\scriptsize\textcolor{green!50!black}{(+0.3)}} & 42.2\,{\scriptsize\textcolor{red!70!black}{($-$0.1)}} \\
 & \quad +\textsc{wHyVol} & 75.9\,{\scriptsize\textcolor{green!50!black}{(+0.9)}} & 67.3\,{\scriptsize\textcolor{red!70!black}{($-$0.1)}} & 58.5\,{\scriptsize\textcolor{gray}{(0.0)}} & 50.2\,{\scriptsize\textcolor{red!70!black}{($-$0.4)}} & 44.9\,{\scriptsize\textcolor{red!70!black}{($-$0.3)}} &  & 76.2\,{\scriptsize\textcolor{green!50!black}{(+3.1)}} & 66.4\,{\scriptsize\textcolor{green!50!black}{(+0.8)}} & 56.5\,{\scriptsize\textcolor{green!50!black}{(+0.1)}} & 47.6\,{\scriptsize\textcolor{red!70!black}{($-$0.1)}} & 41.6\,{\scriptsize\textcolor{red!70!black}{($-$0.7)}} \\
 & \quad +\textsc{wRHG} & 76.6\,{\scriptsize\textcolor{green!50!black}{(+1.6)}} & 67.8\,{\scriptsize\textcolor{green!50!black}{(+0.4)}} & 59.5\,{\scriptsize\textcolor{green!50!black}{(+1.0)}} & 50.7\,{\scriptsize\textcolor{green!50!black}{(+0.1)}} & 45.2\,{\scriptsize\textcolor{gray}{(0.0)}} &  & 76.0\,{\scriptsize\textcolor{green!50!black}{(+2.9)}} & 66.2\,{\scriptsize\textcolor{green!50!black}{(+0.6)}} & 56.5\,{\scriptsize\textcolor{green!50!black}{(+0.1)}} & 47.6\,{\scriptsize\textcolor{red!70!black}{($-$0.1)}} & 41.5\,{\scriptsize\textcolor{red!70!black}{($-$0.8)}} \\
\midrule
FLARE~\cite{flare2026} & \textsc{Gram}$^\star$ & 46.0 & 41.6 & 37.2 & 32.0 & 30.0 &  & 46.8 & 41.6 & 35.9 & 29.4 & 25.9 \\
(T--VA) & \quad +\textsc{HyVol} & 52.0\,{\scriptsize\textcolor{green!50!black}{(+6.0)}} & 47.6\,{\scriptsize\textcolor{green!50!black}{(+6.0)}} & 41.9\,{\scriptsize\textcolor{green!50!black}{(+4.7)}} & 36.3\,{\scriptsize\textcolor{green!50!black}{(+4.3)}} & 33.4\,{\scriptsize\textcolor{green!50!black}{(+3.4)}} &  & 55.5\,{\scriptsize\textcolor{green!50!black}{(+8.7)}} & 48.6\,{\scriptsize\textcolor{green!50!black}{(+7.0)}} & 40.6\,{\scriptsize\textcolor{green!50!black}{(+4.7)}} & 33.2\,{\scriptsize\textcolor{green!50!black}{(+3.8)}} & 28.8\,{\scriptsize\textcolor{green!50!black}{(+2.9)}} \\
 & \quad +\textsc{wHyVol} & 51.7\,{\scriptsize\textcolor{green!50!black}{(+5.7)}} & 46.5\,{\scriptsize\textcolor{green!50!black}{(+4.9)}} & 39.7\,{\scriptsize\textcolor{green!50!black}{(+2.5)}} & 34.9\,{\scriptsize\textcolor{green!50!black}{(+2.9)}} & 32.4\,{\scriptsize\textcolor{green!50!black}{(+2.4)}} &  & 54.6\,{\scriptsize\textcolor{green!50!black}{(+7.8)}} & 47.7\,{\scriptsize\textcolor{green!50!black}{(+6.1)}} & 40.5\,{\scriptsize\textcolor{green!50!black}{(+4.6)}} & 33.1\,{\scriptsize\textcolor{green!50!black}{(+3.7)}} & 28.9\,{\scriptsize\textcolor{green!50!black}{(+3.0)}} \\
 & \quad +\textsc{wRHG} & 53.9\,{\scriptsize\textcolor{green!50!black}{(+7.9)}} & 48.3\,{\scriptsize\textcolor{green!50!black}{(+6.7)}} & 41.9\,{\scriptsize\textcolor{green!50!black}{(+4.7)}} & 36.5\,{\scriptsize\textcolor{green!50!black}{(+4.5)}} & 33.8\,{\scriptsize\textcolor{green!50!black}{(+3.8)}} &  & 55.0\,{\scriptsize\textcolor{green!50!black}{(+8.2)}} & 48.1\,{\scriptsize\textcolor{green!50!black}{(+6.5)}} & 40.7\,{\scriptsize\textcolor{green!50!black}{(+4.8)}} & 33.0\,{\scriptsize\textcolor{green!50!black}{(+3.6)}} & 29.1\,{\scriptsize\textcolor{green!50!black}{(+3.2)}} \\
\cmidrule(l){2-13}
 & \textsc{HyperGram}$^\dag$ & 52.9 & 49.7 & 43.9 & 38.2 & 33.6 &  & 52.6 & 48.3 & 41.4 & 34.4 & 28.6 \\
 & \quad +\textsc{HyVol} & 52.1\,{\scriptsize\textcolor{red!70!black}{($-$0.8)}} & 47.6\,{\scriptsize\textcolor{red!70!black}{($-$2.1)}} & 42.4\,{\scriptsize\textcolor{red!70!black}{($-$1.5)}} & 36.7\,{\scriptsize\textcolor{red!70!black}{($-$1.5)}} & 34.1\,{\scriptsize\textcolor{green!50!black}{(+0.5)}} &  & 56.4\,{\scriptsize\textcolor{green!50!black}{(+3.8)}} & 49.8\,{\scriptsize\textcolor{green!50!black}{(+1.5)}} & 41.5\,{\scriptsize\textcolor{green!50!black}{(+0.1)}} & 33.7\,{\scriptsize\textcolor{red!70!black}{($-$0.7)}} & 29.4\,{\scriptsize\textcolor{green!50!black}{(+0.8)}} \\
 & \quad +\textsc{wHyVol} & 52.1\,{\scriptsize\textcolor{red!70!black}{($-$0.8)}} & 47.4\,{\scriptsize\textcolor{red!70!black}{($-$2.3)}} & 41.4\,{\scriptsize\textcolor{red!70!black}{($-$2.5)}} & 36.2\,{\scriptsize\textcolor{red!70!black}{($-$2.0)}} & 33.7\,{\scriptsize\textcolor{green!50!black}{(+0.1)}} &  & 55.9\,{\scriptsize\textcolor{green!50!black}{(+3.3)}} & 48.7\,{\scriptsize\textcolor{green!50!black}{(+0.4)}} & 41.1\,{\scriptsize\textcolor{red!70!black}{($-$0.3)}} & 33.2\,{\scriptsize\textcolor{red!70!black}{($-$1.2)}} & 29.2\,{\scriptsize\textcolor{green!50!black}{(+0.6)}} \\
 & \quad +\textsc{wRHG} & 52.8\,{\scriptsize\textcolor{red!70!black}{($-$0.1)}} & 47.9\,{\scriptsize\textcolor{red!70!black}{($-$1.8)}} & 42.7\,{\scriptsize\textcolor{red!70!black}{($-$1.2)}} & 37.3\,{\scriptsize\textcolor{red!70!black}{($-$0.9)}} & 34.1\,{\scriptsize\textcolor{green!50!black}{(+0.5)}} &  & 55.8\,{\scriptsize\textcolor{green!50!black}{(+3.2)}} & 48.7\,{\scriptsize\textcolor{green!50!black}{(+0.4)}} & 40.8\,{\scriptsize\textcolor{red!70!black}{($-$0.6)}} & 33.1\,{\scriptsize\textcolor{red!70!black}{($-$1.3)}} & 29.1\,{\scriptsize\textcolor{green!50!black}{(+0.5)}} \\
\midrule
VALOR-32K~\cite{liu2024valor} & \textsc{Gram}$^\star$ & 75.2 & 65.6 & 56.1 & 47.2 & 41.9 &  & 71.2 & 63.0 & 54.0 & 45.4 & 40.4 \\
(T--VA) & \quad +\textsc{HyVol} & 80.5\,{\scriptsize\textcolor{green!50!black}{(+5.3)}} & 71.2\,{\scriptsize\textcolor{green!50!black}{(+5.6)}} & 62.2\,{\scriptsize\textcolor{green!50!black}{(+6.1)}} & 53.3\,{\scriptsize\textcolor{green!50!black}{(+6.1)}} & 47.4\,{\scriptsize\textcolor{green!50!black}{(+5.5)}} &  & 81.6\,{\scriptsize\textcolor{green!50!black}{(+10.4)}} & 72.0\,{\scriptsize\textcolor{green!50!black}{(+9.0)}} & 62.0\,{\scriptsize\textcolor{green!50!black}{(+8.0)}} & 52.0\,{\scriptsize\textcolor{green!50!black}{(+6.6)}} & 46.6\,{\scriptsize\textcolor{green!50!black}{(+6.2)}} \\
 & \quad +\textsc{wHyVol} & 80.4\,{\scriptsize\textcolor{green!50!black}{(+5.2)}} & 70.9\,{\scriptsize\textcolor{green!50!black}{(+5.3)}} & 61.2\,{\scriptsize\textcolor{green!50!black}{(+5.1)}} & 51.8\,{\scriptsize\textcolor{green!50!black}{(+4.6)}} & 47.5\,{\scriptsize\textcolor{green!50!black}{(+5.6)}} &  & 81.1\,{\scriptsize\textcolor{green!50!black}{(+9.9)}} & 71.3\,{\scriptsize\textcolor{green!50!black}{(+8.3)}} & 62.2\,{\scriptsize\textcolor{green!50!black}{(+8.2)}} & 51.6\,{\scriptsize\textcolor{green!50!black}{(+6.2)}} & 46.4\,{\scriptsize\textcolor{green!50!black}{(+6.0)}} \\
 & \quad +\textsc{wRHG} & 80.8\,{\scriptsize\textcolor{green!50!black}{(+5.6)}} & 71.4\,{\scriptsize\textcolor{green!50!black}{(+5.8)}} & 61.7\,{\scriptsize\textcolor{green!50!black}{(+5.6)}} & 53.3\,{\scriptsize\textcolor{green!50!black}{(+6.1)}} & 48.0\,{\scriptsize\textcolor{green!50!black}{(+6.1)}} &  & 80.8\,{\scriptsize\textcolor{green!50!black}{(+9.6)}} & 72.1\,{\scriptsize\textcolor{green!50!black}{(+9.1)}} & 62.8\,{\scriptsize\textcolor{green!50!black}{(+8.8)}} & 52.4\,{\scriptsize\textcolor{green!50!black}{(+7.0)}} & 47.0\,{\scriptsize\textcolor{green!50!black}{(+6.6)}} \\
\cmidrule(l){2-13}
 & \textsc{HyperGram}$^\dag$ & 75.9 & 67.2 & 57.6 & 48.2 & 43.5 &  & 72.4 & 65.1 & 56.2 & 47.3 & 42.1 \\
 & \quad +\textsc{HyVol} & 80.9\,{\scriptsize\textcolor{green!50!black}{(+5.0)}} & 71.3\,{\scriptsize\textcolor{green!50!black}{(+4.1)}} & 62.5\,{\scriptsize\textcolor{green!50!black}{(+4.9)}} & 53.5\,{\scriptsize\textcolor{green!50!black}{(+5.3)}} & 47.6\,{\scriptsize\textcolor{green!50!black}{(+4.1)}} &  & 82.2\,{\scriptsize\textcolor{green!50!black}{(+9.8)}} & 72.5\,{\scriptsize\textcolor{green!50!black}{(+7.4)}} & 62.5\,{\scriptsize\textcolor{green!50!black}{(+6.3)}} & 52.6\,{\scriptsize\textcolor{green!50!black}{(+5.3)}} & 46.9\,{\scriptsize\textcolor{green!50!black}{(+4.8)}} \\
 & \quad +\textsc{wHyVol} & 80.5\,{\scriptsize\textcolor{green!50!black}{(+4.6)}} & 71.0\,{\scriptsize\textcolor{green!50!black}{(+3.8)}} & 60.9\,{\scriptsize\textcolor{green!50!black}{(+3.3)}} & 51.5\,{\scriptsize\textcolor{green!50!black}{(+3.3)}} & 45.5\,{\scriptsize\textcolor{green!50!black}{(+2.0)}} &  & 82.0\,{\scriptsize\textcolor{green!50!black}{(+9.6)}} & 72.2\,{\scriptsize\textcolor{green!50!black}{(+7.1)}} & 62.6\,{\scriptsize\textcolor{green!50!black}{(+6.4)}} & 52.1\,{\scriptsize\textcolor{green!50!black}{(+4.8)}} & 46.6\,{\scriptsize\textcolor{green!50!black}{(+4.5)}} \\
 & \quad +\textsc{wRHG} & 80.9\,{\scriptsize\textcolor{green!50!black}{(+5.0)}} & 71.5\,{\scriptsize\textcolor{green!50!black}{(+4.3)}} & 62.2\,{\scriptsize\textcolor{green!50!black}{(+4.6)}} & 52.6\,{\scriptsize\textcolor{green!50!black}{(+4.4)}} & 46.7\,{\scriptsize\textcolor{green!50!black}{(+3.2)}} &  & 81.8\,{\scriptsize\textcolor{green!50!black}{(+9.4)}} & 72.4\,{\scriptsize\textcolor{green!50!black}{(+7.3)}} & 62.9\,{\scriptsize\textcolor{green!50!black}{(+6.7)}} & 52.5\,{\scriptsize\textcolor{green!50!black}{(+5.2)}} & 47.1\,{\scriptsize\textcolor{green!50!black}{(+5.0)}} \\
\bottomrule
\end{tabular}}
\end{table*}

\clearpage
\end{document}